%% file: main.tex
\documentclass{article}
\usepackage{iclr2026_conference,times}

\input{math_commands.tex}

\usepackage{microtype}
\usepackage{graphicx}
\usepackage{subcaption}
\usepackage{booktabs} 
\usepackage{float}    
\usepackage{multirow}
\usepackage{pdflscape}

\usepackage{hyperref}
\usepackage{url}

\usepackage{amsmath}
\usepackage{amssymb}
\usepackage{mathtools}
\usepackage{amsthm}

\usepackage[capitalize,noabbrev]{cleveref}
\usepackage{enumitem}

\usepackage[disable,textsize=tiny]{todonotes}

\newcommand{\propella}{propella-1}
\newcommand{\propellaann}{propella-annotations}

\title{propella-1: Multi-Property Document Annotation for LLM Data Curation at Scale}

\preprintcopy 

\author{%
\begin{tabular}[t]{c}
\bf Maximilian Idahl\textsuperscript{1,2} \quad Benedikt Droste\textsuperscript{1} \quad Bj\"orn Pl\"uster\textsuperscript{1} \quad Jan Philipp Harries\textsuperscript{1} \\[16pt]
{\bf \textsuperscript{1}ellamind \qquad \textsuperscript{2}Leibniz University Hannover}
\end{tabular}%
}

\begin{document}

\maketitle

\begin{abstract}
Since FineWeb-Edu, data curation for LLM pretraining has predominantly relied on single scalar quality scores produced by small classifiers.
A single score conflates multiple quality dimensions, prevents flexible filtering, and offers no interpretability.
We introduce \propella{}, a family of small multilingual LLMs (0.6B, 1.7B, 4B parameters) that annotate text documents across 18 properties organized into six categories: core content, classification, quality and value, audience and purpose, safety and compliance, and geographic relevance.
The models support 57 languages and produce structured JSON annotations conforming to a predefined schema.
Evaluated against a frontier commercial LLM as a reference annotator, the 4B model achieves higher agreement than much larger general-purpose models.
We release \propellaann{}, a dataset of over three billion document annotations covering major pretraining corpora including data from FineWeb-2, FinePDFs, HPLT 3.0, and Nemotron-CC.
Using these annotations, we present a multi-dimensional compositional analysis of widely used pretraining datasets, revealing substantial differences in quality, reasoning depth, and content composition that single-score approaches cannot capture.
All model weights and annotations are released under permissive, commercial-use licenses.\footnote{Model weights under Apache 2.0, annotations under CC-BY-4.0. All artifacts are available at \url{https://hf.co/collections/ellamind/propella-1}.}
\end{abstract}


\section{Introduction}
\label{sec:introduction}

Data quality is among the highest-leverage factors in LLM pretraining.
FineWeb-Edu matches the next-best open dataset on MMLU with approximately 8$\times$ fewer training tokens through educational quality filtering~\citep{fineweb}, OLMo~3 matches Qwen-3 32B with roughly 6$\times$ fewer tokens through rigorous data curation~\citep{olmo3}, and BeyondWeb reports 7.7$\times$ faster training through systematic document rephrasing~\citep{beyondweb}.
These results establish that investing in data quality yields multiplicative, not marginal, improvements in training efficiency.

The dominant paradigm for model-based data quality filtering since 2024 has been the FineWeb-Edu pipeline: use a large LLM to annotate a sample of documents with a single scalar ``educational value'' score, distill those labels into a small classifier (FastText or a linear head on encoder embeddings), and threshold at scale~\citep{fineweb, li2024dclm}.
This approach has two fundamental limitations.

First, a single scalar score conflates multiple quality dimensions.
Educational value is not the only relevant dimension for pretraining data.
Content with high reasoning depth but low pedagogical structure (e.g., legal analysis, technical specifications) may receive low educational scores despite being highly valuable for training.
Conversely, a document may score well on educational value while containing heavy commercial bias, outdated information, or personally identifiable information.
FineWeb-Edu's own analysis reveals significant topic bias: filtering for educational value systematically favors certain topics (education, history, science) and removes others (entertainment, business, travel)~\citep{fineweb}.

Second, most model-based quality filtering has been developed for English only.
FineWeb-Edu, DCLM, and Nemotron-CC are English-only systems~\citep{fineweb, li2024dclm, nemotroncc}.
While recent work has extended quality filtering to multiple languages through per-language FastText classifiers~\citep{fineweb2hq} or multilingual LLM-as-judge approaches~\citep{jql}, these still rely on single-score quality measures.

We introduce \propella{}, a family of small multilingual LLMs that move beyond single-score filtering to structured multi-property annotation.
Our contributions are as follows:

\begin{enumerate}
\item \textbf{Annotation framework.} We define 18 properties across six categories that capture complementary quality dimensions, enabling flexible, composable filtering strategies for LLM training data curation (\S\ref{sec:framework}).
\item \textbf{propella-1 models.} We train three decoder-only models (0.6B, 1.7B, 4B) based on Qwen-3, that annotate documents in 57 languages, achieving high agreement with frontier models at orders-of-magnitude lower cost (\S\ref{sec:models}, \S\ref{sec:evaluation}).
\item \textbf{propella-annotations.} We release over three billion document annotations covering major pretraining corpora, the largest open release of multi-property annotations for LLM training data to date (\S\ref{sec:scale}).
\item \textbf{Dataset insights.} We demonstrate the analytical value of multi-property annotation through three case studies that reveal compositional differences across datasets, quality tiers, and languages that single-score approaches cannot capture (\S\ref{sec:insights}).
\end{enumerate}


\section{Related Work}
\label{sec:related}

\paragraph{Heuristic filtering.}
Rule-based filtering has been a cornerstone of web-scale pretraining data curation~\citep{oscar, raffel2020c4}.
Common heuristics include length filters, character ratio thresholds, repetition removal, and blocklists.
FineWeb's ablation study reveals that many heuristics have low or even negative impact, and that some C4 rules hurt downstream performance~\citep{fineweb}.
Moreover, \citet{nemotroncc} find that standard heuristic filtering removes 18\% of high-quality tokens, and that disabling heuristic filters for high-quality documents improves downstream accuracy.
Heuristic approaches also carry language-specific assumptions: rules requiring terminal punctuation fail for languages without sentence-final punctuation, and word-length heuristics do not apply to languages without spaces.

\paragraph{Perplexity-based filtering.}
CCNet introduced perplexity filtering using n-gram language models trained on Wikipedia~\citep{ccnet}.
While effective for removing disfluent text across languages, perplexity filtering is biased toward Wikipedia-like content and cannot capture semantic quality: fluent nonsense scores well, while valuable but unconventional content (code, conversational text, technical writing) is penalized~\citep{gopher, fineweb}.

\paragraph{Classifier-based quality scoring.}
Two paradigms have emerged for training quality classifiers.
The first uses curated positive examples (Wikipedia, textbooks) paired with random web samples to train binary classifiers~\citep{li2024dclm, fasttext}.
The second uses LLM-as-judge annotations to train regressors on encoder embeddings, as pioneered by FineWeb-Edu~\citep{fineweb}.
Both produce single scalar quality scores.

\paragraph{Multilingual quality filtering.}
FineWeb-2-HQ trains separate FastText and MLP scorers per language using diverse multilingual training sources~\citep{fineweb2hq}.
JQL evaluates LLM judges for educational quality across 35 languages and distills their scores into lightweight regressors~\citep{jql}.
These extend single-score filtering to non-English languages but do not address the single-score limitation itself.

\paragraph{Register-based annotation.}
Recent work on web register classification annotates documents with hierarchical register labels (e.g., narrative, informational, instructional) rather than a single quality score~\citep{registerclassification, registermatters}.
HPLT 3.0~\citep{hplt3} applies register annotations at web scale alongside quality and PII labels, representing the broadest annotation effort among these systems.
\citet{registermatters} show that register-based data selection can improve or harm LLM performance, indicating that text type matters independently of quality.

\paragraph{Multi-dimensional quality assessment.}
Several recent systems assess data quality along multiple dimensions rather than a single score.
QuRating~\citep{qurating} annotates a 260B token corpus with scalar ratings across four quality dimensions using pairwise LLM judgments.
DataMan~\citep{dataman} derives 14 quality criteria and annotates 447B tokens, while Meta-rater~\citep{metarater} integrates four quality dimensions with existing signals through learned weightings, demonstrating that multi-dimensional selection can double convergence speed.
RedPajama~V2~\citep{redpajamav2} accompanies over 100 trillion tokens of multilingual web text with quality signals and metadata.
These approaches produce scalar ratings or heuristic signals for data selection; none combines structured categorical annotations, broad property coverage (including safety, geographic relevance, and content classification), multilingual support, and a public release of large-scale annotations.

\medskip

\propella{} addresses this gap with structured categorical annotation across 18 properties for 57 languages, at billion-document scale, with all annotations publicly released.


\section{Annotation Framework}
\label{sec:framework}

\subsection{Design Philosophy}

Our annotation framework is motivated by the observation that data quality is inherently multi-dimensional.
A document's value for LLM pretraining depends on its content integrity, topical relevance, reasoning depth, audience level, safety profile, and many other factors.
Reducing this to a single scalar discards information that practitioners need for flexible, task-specific curation.

We designed the framework to satisfy three requirements:
(1) each property should enable a concrete filtering use case;
(2) properties should be complementary, not redundant;
(3) annotation values should use enumerated categories with clear rubric definitions to ensure consistency.

\subsection{Properties}

\propella{} annotates documents across 18 properties organized into six categories, summarized in Table~\ref{tab:properties}.

Each property was chosen because it enables a concrete curation scenario.
Educational value supports curriculum-style training schedules.
Reasoning indicators allow selecting content rich in analytical and explanatory reasoning, relevant for training reasoning models.
Commercial bias enables filtering out marketing-dominated content that may degrade model objectivity.
Time-sensitivity distinguishes evergreen reference material from ephemeral news, supporting freshness-aware data selection.
Content type and business sector enable domain-targeted dataset construction.
Safety and PII properties support compliance-aware filtering.

Because annotations are structured as typed JSON fields, practitioners can compose arbitrary filtering predicates, for example selecting documents that simultaneously satisfy high reasoning depth, low commercial bias, and no PII.
A representative annotation output is shown in Figure~\ref{fig:json-example} (Appendix~\ref{app:json-example}).

\begin{table}[t]
\centering
\caption{The 18 annotation properties of \propella{}, organized by category. $|V|$ denotes the number of enumerated values; -- indicates open vocabulary (free-text or ISO-3166 country names). Property type indicates the evaluation metric used: ordinal properties are evaluated with Quadratic Weighted Kappa (QWK), the binary property with F1, multi-select properties with Intersection-over-Union (IoU, also known as the Jaccard index), and the free-text property is excluded from quantitative evaluation.}
\label{tab:properties}
\small
\begin{tabular}{@{}llllr@{}}
\toprule
\textbf{Category} & \textbf{Property} & \textbf{Description} & \textbf{Type} & $|V|$ \\
\midrule
\multirow{3}{*}{Core Content} & Content Integrity & Completeness and technical quality & Ordinal & 4 \\
 & Content Ratio & Content vs.\ navigation/UI ratio & Ordinal & 5 \\
 & Content Length & Amount of substantive content & Ordinal & 4 \\
\midrule
\multirow{4}{*}{Classification} & One-Sentence Desc. & Neutral summary of the document & Free-text & -- \\
 & Content Type & Functional structure and purpose & Multi-select & 18 \\
 & Business Sector & Industry domain relevance & Multi-select & 37 \\
 & Technical Content & Specialized knowledge type & Multi-select & 7 \\
\midrule
\multirow{4}{*}{Quality \& Value} & Content Quality & Writing and presentation quality & Ordinal & 5 \\
 & Information Density & Signal vs.\ redundancy ratio & Ordinal & 5 \\
 & Educational Value & Teaching and learning potential & Ordinal & 5 \\
 & Reasoning Indicators & Logical reasoning depth & Ordinal & 5 \\
\midrule
\multirow{3}{*}{Audience \& Purpose} & Audience Level & Target sophistication level & Ordinal & 6 \\
 & Commercial Bias & Commercial influence on objectivity & Ordinal & 5 \\
 & Time-Sensitivity & Temporal value decay & Ordinal & 4 \\
\midrule
\multirow{2}{*}{Safety} & Content Safety & Inappropriate or harmful content & Ordinal & 5 \\
 & PII Presence & Personally identifiable information & Binary & 2 \\
\midrule
\multirow{2}{*}{Geographic} & Regional Relevance & Regional/cultural context & Multi-select & 14 \\
 & Country Relevance & Specific country relevance & Multi-select & -- \\
\bottomrule
\end{tabular}
\end{table}

\subsection{Rubric Development}

Consistent annotation at scale requires precise operational definitions.
We developed the annotation rubric through a systematic iterative process over approximately two weeks.
Starting from a large set of candidate properties drawn from data curation practice, we consolidated overlapping dimensions and removed properties that did not enable distinct filtering use cases, arriving at a core set of 17.
Through subsequent pilot annotation rounds, we refined definitions, adjusted category boundaries, and resolved ambiguities.
The one-sentence description property was added late in the process to enable English-speaking practitioners to inspect and navigate multilingual datasets without requiring knowledge of the source language, yielding the final set of 18.

The complete rubric spans approximately 8,000 words and constitutes a comprehensive annotation manual.
For each property, the rubric specifies all permissible values with detailed selection criteria, concrete positive and negative examples, and key decision points that disambiguate between adjacent categories.
For ordinal properties such as content quality and content integrity, percentage-based thresholds and graduated severity levels provide annotators (and models) with clear boundaries.
Multi-select properties such as content type (18 possible values), business sector (37 sectors), and technical content (7 categories) include multi-label combination examples that illustrate when and how to assign multiple labels simultaneously.

Several properties receive particularly detailed treatment.
Educational value and reasoning indicators include structured decision trees, automation heuristics (e.g., lexical and structural cue counts per thousand tokens), disambiguation rules that distinguish related concepts (e.g., explanatory vs.\ analytical reasoning, procedural steps vs.\ logical argumentation), and annotated borderline examples.
The information density property defines common repetition patterns to watch for, such as copy-paste artifacts and circular reasoning.
Safety and PII properties include explicit guidelines for distinguishing contextual mentions (e.g., public figures, academic authors) from genuine privacy concerns.

The rubric also addresses challenges specific to multilingual annotation.
Language-specific quality indicators account for proper character encoding in non-Latin scripts, adjusted word-density expectations for agglutinative languages such as Turkish and Finnish, and appropriate register assessment for languages with formalized politeness systems such as Japanese and Korean.
Regional and country relevance properties include explicit guidelines clarifying that language does not determine geographic relevance (e.g., Spanish-language content about Asian markets should not be classified as Latin American).

The full rubric is provided in Appendix~\ref{app:properties} and released alongside the model weights.


\section{propella-1 Models}
\label{sec:models}

\subsection{Architecture}

The \propella{} family consists of three models based on the Qwen-3 architecture~\citep{qwen3}: 0.6B, 1.7B, and 4B parameters.
We chose decoder-only models over encoder-based classifiers for three reasons: (1) they handle long documents natively (up to 64K context), (2) they produce all 18 annotations as structured JSON output removing the need for per-property classifier training, and (3) through fine-tuning, they internalize detailed annotation guidelines and achieve strong performance with a short, efficient prompt.

\subsection{Training Data}

Training data was obtained by annotating a diverse document sample using multiple frontier LLMs, prompted with the full annotation rubric and strict response schema.
The training set covers 57 languages with approximately 35\% English and the remainder spanning European languages, Arabic, Chinese, Japanese, Korean, Thai, and others.
Dedicated support for code, mathematical content, and post-training data (instruction-following, conversations) is included.
Documents were sampled from diverse sources including web crawls, PDFs, and curated datasets to ensure the models generalize across content types.
Detailed breakdowns of the language and source distributions are provided in Appendix~\ref{app:training-data}.

To address limitations of API content filters, which sometimes refused to annotate problematic documents, a small subset of documents was annotated manually.

\subsection{Training Setup}

Models were fine-tuned with a 64K context length using fp8 mixed-precision training on 4$\times$ H100 GPUs within a few hours per model variant.
The target output format is a JSON object conforming to a predefined schema with enumerated values for all categorical properties.
No whitespace is included in the JSON output, to reduce the number of output tokens during inference.

\subsection{Prompt Design}

The \propella{} models use a compact system prompt (${\sim}800$ tokens) that lists all properties with their enumerated values and brief descriptions.
This prompt is significantly shorter than the full annotation prompt (${\sim}14$K tokens) used during training data creation.


\section{Evaluation}
\label{sec:evaluation}

\subsection{Setup}

We evaluate on a test set of 3,000 documents.
For these documents, we obtain annotations from Gemini-3-Pro with reasoning effort set to ``high''.
We use these as reference labels, treating Gemini-3-Pro as a high-quality proxy for this structured annotation task.
We discuss the implications and limitations of this choice in \S\ref{sec:discussion}.
The one-sentence description property is excluded from quantitative evaluation.

\subsection{Metrics}

Properties are grouped by type, each evaluated with an appropriate metric:

\begin{itemize}
\item \textbf{Ordinal properties} (11 properties): Quadratic Weighted Kappa (QWK), which measures agreement while accounting for the ordinal nature of labels, penalizing larger disagreements more heavily.
\item \textbf{Binary property} (1 property: PII presence): F1 score.
\item \textbf{Multi-select properties} (5 properties): Intersection-over-Union (IoU), averaged across samples.
\end{itemize}

We compute an overall score as a weighted average reflecting the proportion of each property type:
\begin{equation}
\text{overall} = \frac{11}{17} \times \overline{\text{QWK}} + \frac{1}{17} \times \text{F1} + \frac{5}{17} \times \overline{\text{IoU}}
\end{equation}

\subsection{Baselines}

We compare against several general-purpose LLMs, including models from the Gemini~\citep{gemini25}, Qwen~\citep{qwen3}, and Gemma~\citep{gemma3} families, as well as Mistral-Small~\citep{mistralsmall} and SmolLM3~\citep{smollm3}.
All baseline models use the detailed annotator system prompt, which includes the complete rubric with examples, decision criteria, and language-specific guidelines.
The \propella{} models use their compact prompt.
We also evaluated some baseline models with the compact \propella{} prompt, which consistently led to worse performance, as the short prompt lacks the detail needed for models not specifically trained for this task.

\subsection{Results}

\begin{figure}[t]
    \centering
    \includegraphics[width=0.90\textwidth]{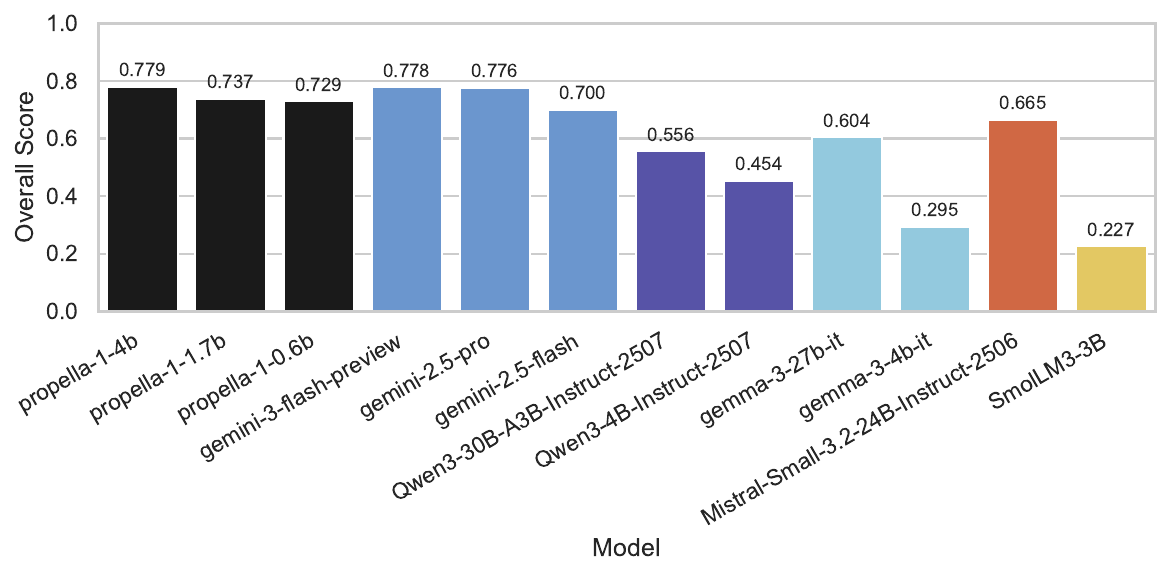}
    \caption{Overall annotation agreement scores across all evaluated models. propella-1-4b exceeds Gemini-3-Flash and significantly larger open models.}
    \label{fig:overall}
    \end{figure}

\begin{figure}[t]
    \centering
    \includegraphics[width=\textwidth]{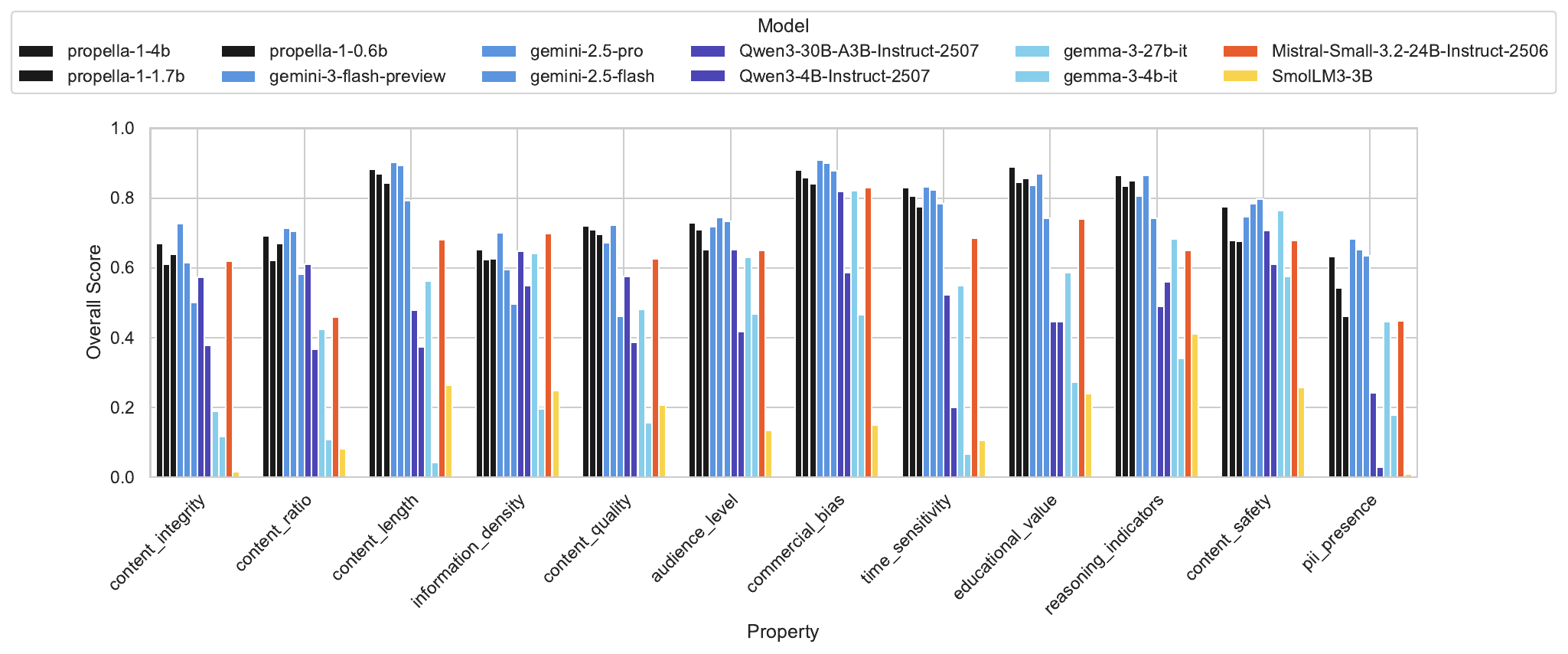}
    \caption{Per-property annotation agreement scores across all evaluated models (12 properties). See Figure~\ref{fig:perproperty_full} in Appendix~\ref{app:full_results} for the full breakdown of all 17 properties.}
    \label{fig:perproperty}
\end{figure}

Figure~\ref{fig:overall} shows the overall performance comparison across all evaluated models.
The 4B model achieves an overall score of 0.779, exceeding Gemini-3-Flash and all open-weight baselines despite being significantly smaller.
Even the 0.6B model achieves a strong score of 0.729, demonstrating that small, specialized models can approach the annotation quality of much larger general-purpose models on this task.
Figure~\ref{fig:perproperty} presents the per-property score breakdown, revealing that \propella{} models perform well across all properties.
We also find that fp8 inference preserves annotation quality with negligible differences compared to bf16 (see Appendix~\ref{app:fp8}).

\section{Annotation at Scale}
\label{sec:scale}

\subsection{Serving Infrastructure}

We serve \propella{} models using SGLang~\citep{sglang} with the llguidance structured output backend~\citep{llguidance}, enforcing strict JSON schema conformance and no whitespace during generation, guaranteeing that every output is a valid annotation object with correct enumerated values.
This eliminates the need for response schema validation or retry logic.

Table~\ref{tab:throughput} reports throughput benchmarks.
After a short warmup phase, we run inference for 5,000 documents, sending 1,000 concurrent requests to the SGLang server, to obtain a rough estimate of the GPU-hours required to annotate one million documents.
On a single H100 GPU with fp8 precision, the 4B model processes 27.0 documents per second, corresponding to approximately 10.3 GPU-hours per million documents.
Since annotation inputs are full documents while outputs are compact JSON objects, throughput is dominated by prefill: the 4B fp8 configuration achieves 50.1K prompt tokens per second versus 3.9K output tokens per second.

\begin{table}[t]
\centering
\caption{Inference throughput in documents per second and tokens-per-second (TPS) for \propella{} models. GPU-hours per 1M documents estimates the compute cost for large-scale annotation. Throughput is dominated by prefill, as annotation inputs are full documents while outputs are compact JSON objects.}
\label{tab:throughput}
\begin{tabular}{@{}llcccc@{}}
\toprule
\textbf{Model} & \textbf{GPU} & \textbf{Docs/s} & \textbf{h / 1M docs} & \textbf{Prompt TPS} & \textbf{Output TPS} \\
\midrule
propella-1-4b & A100 80GB & 10.3 & 27.0 & 19.1K & 1.5K \\
propella-1-4b & H100 96GB & 22.4 & 12.4 & 41.6K & 3.2K \\
propella-1-4b (fp8) & H100 96GB & 27.0 & 10.3 & 50.1K & 3.9K \\
propella-1-1.7b & A100 80GB & 17.8 & 15.6 & 33.0K & 2.6K \\
propella-1-1.7b & H100 96GB & 35.8 & 7.8 & 66.5K & 5.2K \\
propella-1-1.7b (fp8) & H100 96GB & 39.1 & 7.1 & 72.7K & 5.7K \\
propella-1-0.6b & A100 80GB & 21.5 & 12.9 & 40.0K & 3.1K \\
propella-1-0.6b & H100 96GB & 39.9 & 7.0 & 74.2K & 5.7K \\
\bottomrule
\end{tabular}
\end{table}

\subsection{Scaling}

For large-scale annotation, we scaled inference across thousands of GPUs using inference-hive,\footnote{\url{https://github.com/ellamind/inference-hive}} an open-source distributed inference orchestration system for SLURM clusters.
In one deployment, we ran the 4B model on 3,936 A100 GPUs and annotated ${\sim}500$ million documents from FineWeb-2 in approximately 3.5 hours.

\subsection{The propella-annotations Dataset}

We release \propellaann{}, a dataset of over three billion document annotations covering major pretraining corpora.
All annotations were produced with the propella-1-4b model using the serving infrastructure described above.
Table~\ref{tab:annotations} summarizes the dataset.

The annotated corpora span a diverse set of pretraining sources: web crawls (FineWeb-2, HPLT~3.0, Nemotron-CC, German Commons), extracted PDFs (FinePDFs), encyclopedic text (finewiki), and synthetic data (SYNTH, a collection of synthetic instruction-following and conversation data).
While the \propella{} models support 57 languages, the current annotation release covers English and 14 European languages, with the largest volumes in German, Spanish, French, and Italian.
The dataset will be extended with additional corpora and languages over time.

Each annotation record contains the full set of 18 property values for a single document, keyed by the document identifier from the source dataset.
This enables direct joining with the original corpora for downstream filtering and selection without redistributing the source text.
Because the annotations consist of categorical values and brief text descriptions rather than full documents, their small storage and memory footprint makes it practical to analyze and filter large pretraining datasets on modest hardware.

The dataset is hosted on the Hugging Face Hub,\footnote{\url{https://hf.co/datasets/openeurollm/propella-annotations}} including a detailed dataset card with usage examples.
All annotations are released under the CC-BY-4.0 license.

\begin{table}[t]
\centering
\caption{Overview of the \propellaann{} dataset. All annotations were produced with propella-1-4b.}
\label{tab:annotations}
\begin{tabular}{@{}llr@{}}
\toprule
\textbf{Source Dataset} & \textbf{Languages} & \textbf{Annotations} \\
\midrule
FineWeb-2 & DE, ES, FR, IT, SV, FI & 1,632,650,735 \\
HPLT 3.0 & DE, FI & 694,920,477 \\
FinePDFs & EN + 14 EU languages & 365,048,869 \\
Nemotron-CC & EN (high-quality split) & 155,688,999 \\
SYNTH & Multilingual & 77,908,583 \\
finewiki & Multilingual & 43,097,138 \\
German Commons & DE & 35,716,016 \\
\midrule
\textbf{Total} & & \textbf{3,005,080,817} \\
\bottomrule
\end{tabular}
\end{table}


\section{Dataset Insights}
\label{sec:insights}

We demonstrate the analytical value of multi-property annotation through three case studies.
These findings are only possible with structured multi-property annotation at scale and illustrate why a single quality score is insufficient for understanding pretraining data.

\subsection{German Multi-Source Profiling}

We compare property distributions across four German-language pretraining sources: FineWeb-2~\citep{fineweb2} (496M docs), FinePDFs~\citep{finepdfs} (36M docs), HPLT 3.0~\citep{hplt3} (645M docs), and German Commons~\citep{germancommons} (36M docs).
By controlling for language, we isolate the effect of data source on content characteristics.

\begin{figure}[t]
\centering
\includegraphics[width=\textwidth]{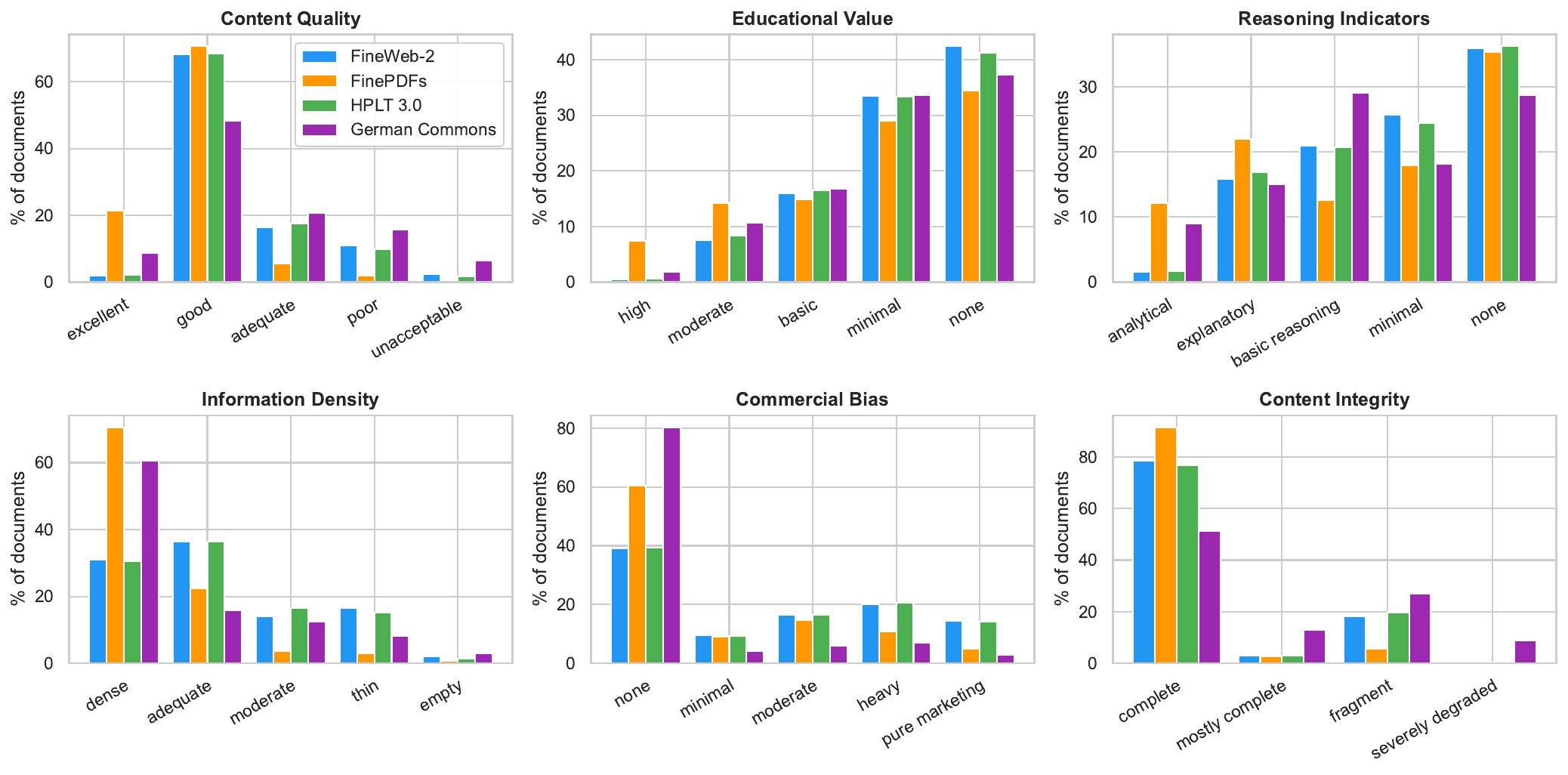}
\caption{Property distributions across four German-language pretraining sources. Sources differ dramatically across quality dimensions despite all being German text corpora. FinePDFs shows substantially higher rates of excellent quality, analytical reasoning, and high educational value.}
\label{fig:german_sources}
\end{figure}

Figure~\ref{fig:german_sources} reveals that these pools of German text have dramatically different quality profiles.
FinePDFs contains substantially more content rated as ``excellent'' quality (21.4\% vs.\ 2.4\% in FineWeb-2), approximately 12$\times$ more content with analytical reasoning indicators, and 7.4\% high educational value compared to 0.6\% in FineWeb-2.
HPLT 3.0, as a less curated web crawl, shows higher rates of content fragments and degraded documents compared to FineWeb-2.

These differences have direct implications for data mixture design: a practitioner seeking reasoning-heavy training data should prioritize FinePDFs, while one optimizing for scale with acceptable quality should consider FineWeb-2.
This information is difficult to capture with single-score quality classifiers.

\subsection{Nemotron-CC Quality Tier Audit}

Nemotron-CC~\citep{nemotroncc} classifies documents into quality tiers using a classifier ensemble.
We analyze 10,000 documents from each of the five quality tiers using \propella{} annotations to audit their multi-dimensional quality profiles.
Specifically, we examine whether documents in the ``high'' quality tier are uniformly high-quality across all dimensions.

\begin{figure}[t]
\centering
\includegraphics[width=0.6\textwidth]{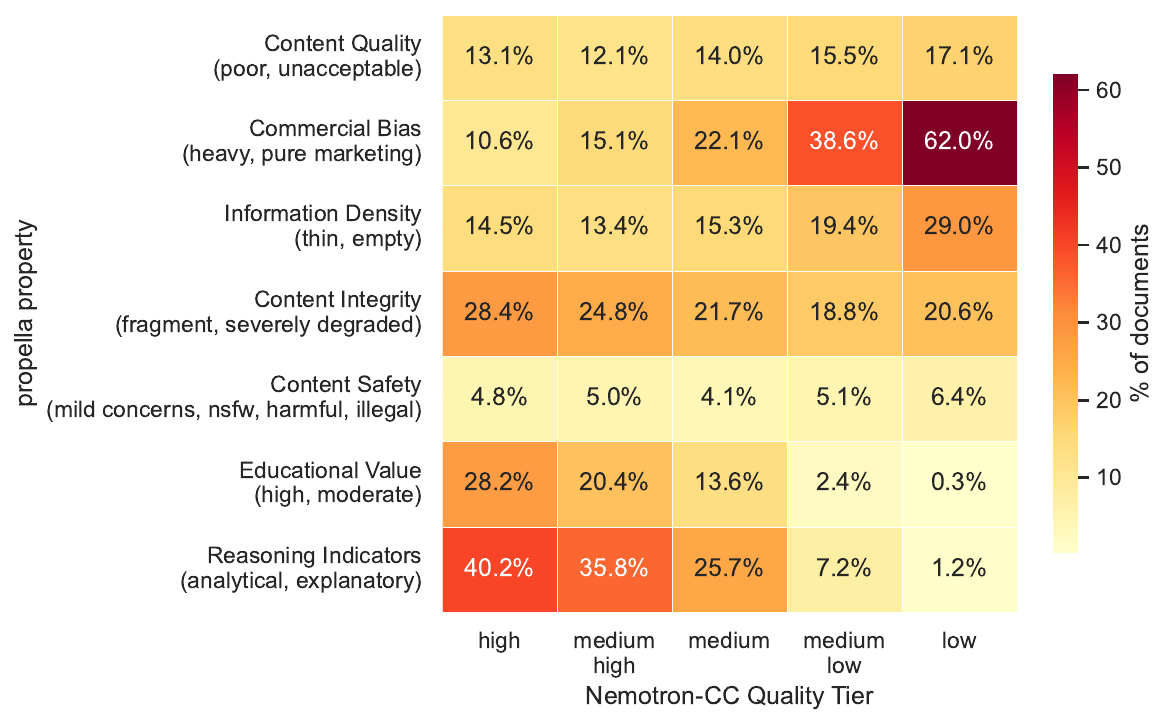}
\caption{Prevalence of quality issues in Nemotron-CC quality tiers. Even the ``high'' quality tier contains documents with issues on specific dimensions (commercial bias, information density, content integrity) that the single-score classifier does not capture.}
\label{fig:nemotron_audit}
\end{figure}

Figure~\ref{fig:nemotron_audit} shows the prevalence of specific quality issues across tiers.
While Nemotron-CC's classifier tiers do correlate with several \propella{} quality dimensions (higher tiers have better content quality and information density on average), the ``high'' tier still contains documents with issues that multi-property annotation reveals.
These include documents with heavy or pure marketing commercial bias, thin information density, and incomplete content integrity, among others.

This finding illustrates a key limitation of single-score quality classification: documents can score well on one aggregate measure while having undesirable characteristics on specific dimensions that matter for particular downstream applications.

\subsection{Cross-Language Content Variation in FineWeb-2}

FineWeb-2 applies the same crawl and processing pipeline across languages~\citep{fineweb2}.
We examine whether this produces comparable quality profiles by comparing property distributions across six languages: German, Spanish, French, Italian, Swedish, and Finnish.

\begin{figure}[t]
\centering
\includegraphics[width=0.7\textwidth]{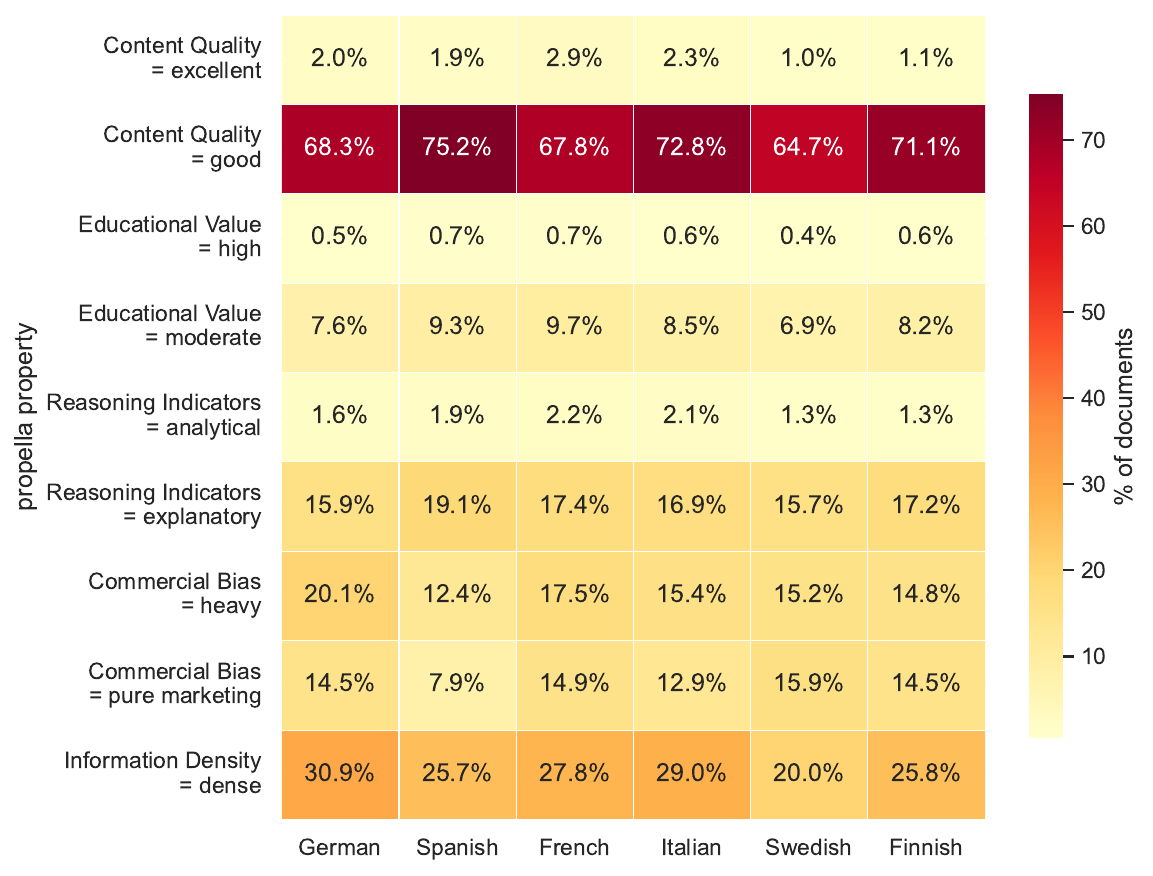}
\caption{Property distributions across six languages in FineWeb-2. The rightmost column shows the range (max $-$ min) across languages. Commercial bias and information density exhibit the largest cross-language variation, while educational value and reasoning indicators are more uniform. These differences motivate language-specific filtering strategies.}
\label{fig:cross_language}
\end{figure}

Figure~\ref{fig:cross_language} shows that the same pipeline yields substantially different content profiles by language.
Languages differ in their distributions of content quality, commercial bias, educational value, and content type composition.
These differences likely reflect the underlying composition of each language's web presence rather than pipeline artifacts.

The implication is that language-specific filtering thresholds are needed for effective multilingual data curation.
A single quality threshold applied uniformly across languages may be too aggressive for some languages and too lenient for others.

\section{Discussion and Future Work}
\label{sec:discussion}

\paragraph{Value of the release.}
The \propellaann{} dataset enables researchers to study which quality dimensions matter for downstream LLM training without incurring the cost of annotation.
By providing pre-computed annotations for over three billion documents across major pretraining corpora, we lower the barrier to data-centric research.

\paragraph{Limitations.}
Our evaluation relies on agreement with Gemini-3-Pro rather than human annotations, which introduces a dependency on a specific model's judgments.
Because the distilled models and the evaluation reference may share systematic biases originating from frontier LLM outputs, high agreement could mask shared errors that human evaluation would reveal.
This risk is compounded by the tendency of LLM evaluators to favor outputs resembling their own generations~\citep{selfpreference}.
Additionally, the \propella{} models inherit biases from the frontier LLMs used to generate training labels, which may affect annotation consistency for underrepresented languages or content types.
LLM-based judges are also known to vary in reliability across languages~\citep{multilingual_llm_as_judge_reliability}, and our per-language evaluation samples may be too small to detect such variation reliably.
The property rubric itself reflects one team's design choices and may not capture all dimensions relevant to every use case.
Most importantly, we have not yet evaluated whether filtering with \propella{} annotations improves downstream model training, which is the ultimate measure of their utility.
However, the case studies in \S\ref{sec:insights} demonstrate the analytical value of multi-property annotation, and the public release of all annotations enables any researchers to conduct downstream training experiments.

\paragraph{Future work.}
The most important next step is evaluating the downstream training impact of multi-property filtering.
Does selecting documents that are high on educational value, reasoning indicators, and content quality simultaneously produce better models than selecting on any single dimension?
Our annotations and their public release make this experiment accessible to any research group.
We plan to perform various ablation experiments to evaluate the impact of different quality dimensions on downstream task performance.
Additional directions include property-aware curriculum learning, where the training data mixture evolves based on multiple quality dimensions over the course of training, and integration with data transformation and synthetic generation pipelines, where multi-property annotations could guide which documents to rephrase, augment, or discard.


\section{Conclusion}
\label{sec:conclusion}

We presented \propella{}, a family of small multilingual LLMs for multi-property document annotation, and \propellaann{}, a large-scale open multi-property annotation of LLM pretraining data covering over three billion documents.
The annotations are accurate (the 4B model exceeds Gemini-3-Flash in agreement with Gemini-3-Pro), efficient to produce (27 documents per second per H100 GPU), highly multilingual (57 languages), and openly available.
Through three analytical case studies, we demonstrated that multi-property annotation reveals compositional structure in pretraining datasets that single-score approaches cannot capture: different data sources have dramatically different quality profiles, single-score quality tiers mask heterogeneity, and the same processing pipeline produces different quality distributions across languages.
We release all model weights and annotations to enable the community to build on this resource for data mixture design, compliance-aware governance, and multilingual data curation research.


\subsubsection*{Acknowledgments}

\begin{itemize}[leftmargin=*,nosep]
\item This project used compute resources made available via the EuroHPC Joint Undertaking (EuroHPC JU) AI Factories initiative (AI for Industrial Innovation -- Large Scale Access) on the EuroHPC supercomputer LEONARDO operated by CINECA and the LEONARDO consortium.
\item This project used compute resources made available via the EuroHPC Joint Undertaking (EuroHPC JU) AI Factories initiative (AI for Industrial Innovation -- Large Scale Access) on the EuroHPC supercomputer MareNostrum~5 operated by the Barcelona Supercomputing Center (BSC).
\item This project is supported by the OpenEuroLLM project, co-funded by the Digital Europe Programme under GA no.\ 101195233.
For more information see \url{https://openeurollm.eu}.
\item This project is supported by the LLMs4EU project, co-funded by the Digital Europe Programme under GA no.\ 101198470.
For more information see the \href{https://www.alt-edic.eu/projects/llms4eu}{LLMs4EU website}.
\item ellamind is supported by the German Federal Ministry for Economic Affairs and Energy (BMWE) under the \href{https://www.soofi.info/}{soofi} (Sovereign Open Source Foundation Models for European Intelligence) project.
\item ellamind thanks the AI Service Center for Sensitive and Critical Infrastructures (\href{https://kisski.gwdg.de}{KISSKI}), operated by GWDG, for additional compute access.
\end{itemize}


\subsubsection*{Ethics Statement}

The \propellaann{} dataset includes safety and PII annotations for over three billion documents.
While these labels are intended to support filtering out unsafe or privacy-sensitive content, they could in principle be misused to locate such content.
We note that the annotations contain only categorical labels and brief text descriptions, not the source documents themselves, which limits this risk.
The safety and PII labels produced by \propella{} are not a substitute for dedicated content moderation systems and should be used as one signal among many in a responsible data curation pipeline.
The annotation rubric reflects one team's design choices and may encode biases, particularly for underrepresented languages and cultural contexts.
The models inherit biases from the frontier LLMs used to generate training labels, and users should be aware of this when applying annotations to sensitive filtering decisions.

\subsubsection*{Reproducibility Statement}

We release all artifacts necessary to reproduce and build upon this work.\footnote{\url{https://huggingface.co/collections/ellamind/propella-1}}
Model weights for all three \propella{} variants (0.6B, 1.7B, 4B) are released under the Apache 2.0 license.
The \propellaann{} dataset of over three billion annotations is hosted on the Hugging Face Hub under CC-BY-4.0.
The complete annotation rubric is provided in Appendix~\ref{app:properties} and released alongside the model weights.
Serving configuration, prompt design, and inference throughput details are described in Sections~\ref{sec:models} and~\ref{sec:scale}, and are included in the model repository.
Training data composition is detailed in Appendix~\ref{app:training-data}.


\bibliography{refs}
\bibliographystyle{iclr2026_conference}


\appendix

\section{Training Data Composition}
\label{app:training-data}

Tables~\ref{tab:language-dist} and~\ref{tab:source-dist} show the language and source distributions of the training data, respectively.

\begin{table}[H]
\centering
\caption{Language distribution of the training data. Language codes follow the FLORES-200 convention: ISO~639-3 language codes combined with ISO~15924 script codes (e.g., \texttt{eng\_Latn} = English in Latin script).}
\label{tab:language-dist}
\begin{tabular}{lr@{\hspace{1.5em}}lr@{\hspace{1.5em}}lr}
\toprule
\textbf{Language} & \textbf{\%} & \textbf{Language} & \textbf{\%} & \textbf{Language} & \textbf{\%} \\
\midrule
eng\_Latn & 35.08 & por\_Latn & 0.90 & glg\_Latn & 0.83 \\
spa\_Latn & 3.98  & kat\_Geor & 0.90 & gle\_Latn & 0.83 \\
ita\_Latn & 3.97  & ben\_Beng & 0.90 & nno\_Latn & 0.83 \\
fra\_Latn & 3.95  & fas\_Arab & 0.89 & ltg\_Latn & 0.77 \\
deu\_Latn & 3.86  & nob\_Latn & 0.86 & yue\_Hant & 0.49\\
pol\_Latn & 3.81  & ekk\_Latn & 0.89 & cmn\_Hant & 0.48 \\
code      & 2.82  & fin\_Latn & 0.89 & hrv\_Latn & 0.43 \\
math      & 2.77  & tur\_Latn & 0.89 & arb\_Arab & 0.39\\
sft       & 2.41  & swe\_Latn & 0.88 & bos\_Latn & 0.39 \\
ukr\_Cyrl & 0.95  & ind\_Latn & 0.88 & mkd\_Cyrl & 0.39 \\
nld\_Latn & 0.95  & ces\_Latn & 0.88 & srp\_Latn & 0.37 \\
tha\_Thai & 0.95  & lit\_Latn & 0.88 & cmn\_Hani & 0.37 \\
jpn\_Jpan & 0.94  & eus\_Latn & 0.87 & hin\_Deva & 0.36 \\
heb\_Hebr & 0.94  & vie\_Latn & 0.87 & srp\_Cyrl & 0.36 \\
ell\_Grek & 0.93  & slv\_Latn & 0.87 & als\_Latn & 0.35 \\
kor\_Hang & 0.93  & mlt\_Latn & 0.86 & sqi\_Latn & 0.03 \\
isl\_Latn & 0.93  & bul\_Cyrl & 0.86 & est\_Latn & 0.02  \\
dan\_Latn & 0.92  & lvs\_Latn & 0.86 & lav\_Latn & 0.02 \\
cat\_Latn & 0.92  & urd\_Arab & 0.85 & nor\_Latn & 0.02 \\
slk\_Latn & 0.92  & hun\_Latn & 0.85 & swa\_Latn & 0.02 \\
rus\_Cyrl & 0.91  & ron\_Latn & 0.84 &           &  \\

\bottomrule
\end{tabular}
\end{table}

\begin{table}[H]
\centering
\caption{Source distribution of the training data.}
\label{tab:source-dist}
\begin{tabular}{lr@{\hspace{2em}}lr}
\toprule
\textbf{Source} & \textbf{\%} & \textbf{Source} & \textbf{\%} \\
\midrule
HPLT 3.0 (unfiltered)          & 39.59 & Nemotron SFT code               & 0.41 \\
FineWeb                        & 16.01 & Nemotron-CC diverse QA          & 0.41 \\
FineWeb-2                      & 13.23 & RedPajama arXiv sample          & 0.40 \\
FinePDFs                       &  8.09 & RedPajama Wikipedia sample      & 0.40 \\
FineWeb-2 (removed)            &  6.28 & Nemotron-CC high-quality synth  & 0.39 \\
FineWeb-Edu (dedup)            &  3.92 & Nemotron-CC translated diverse  & 0.38 \\
The Stack                      &  2.04 & Nemotron synthetic code         & 0.38 \\
FineMath                       &  2.00 & RedPajama GitHub sample         & 0.37 \\
OpenHermes                     &  2.00 & RedPajama StackExchange sample  & 0.36 \\
FineWiki                       &  1.35 & Nemotron-CC math                & 0.35 \\
RedPajama C4 sample            &  0.42 & Nemotron-CC high-quality        & 0.33 \\
Nemotron SFT general           &  0.42 & RedPajama book sample           & 0.07 \\
Nemotron SFT math              &  0.42 &                                 &      \\
\bottomrule
\end{tabular}
\end{table}

\section{Example Annotation Output}
\label{app:json-example}

\begin{figure}[H]
\centering
\small
\begin{verbatim}
{"content_integrity":"complete",
 "content_ratio":"mostly_content",
 "content_length":"substantial",
 "one_sentence_description":"Research paper on ...",
 "content_type":["analytical"],
 "business_sector":["academic_research"],
 "technical_content":["scientific","data_heavy"],
 "content_quality":"excellent",
 "information_density":"dense",
 "educational_value":"high",
 "reasoning_indicators":"analytical",
 "audience_level":"expert",
 "commercial_bias":"none",
 "time_sensitivity":"slowly_changing",
 "content_safety":"safe",
 "pii_presence":"no_pii",
 "regional_relevance":["european"],
 "country_relevance":["germany"]}
\end{verbatim}
\caption{Representative \propella{} annotation output (JSON, formatted for readability). During inference, the output contains no whitespace to minimize token count.}
\label{fig:json-example}
\end{figure}

\section{Full Per-Property Evaluation Results}
\label{app:full_results}

\begin{landscape}
\begin{figure}[p]
    \centering
    \includegraphics[width=\linewidth]{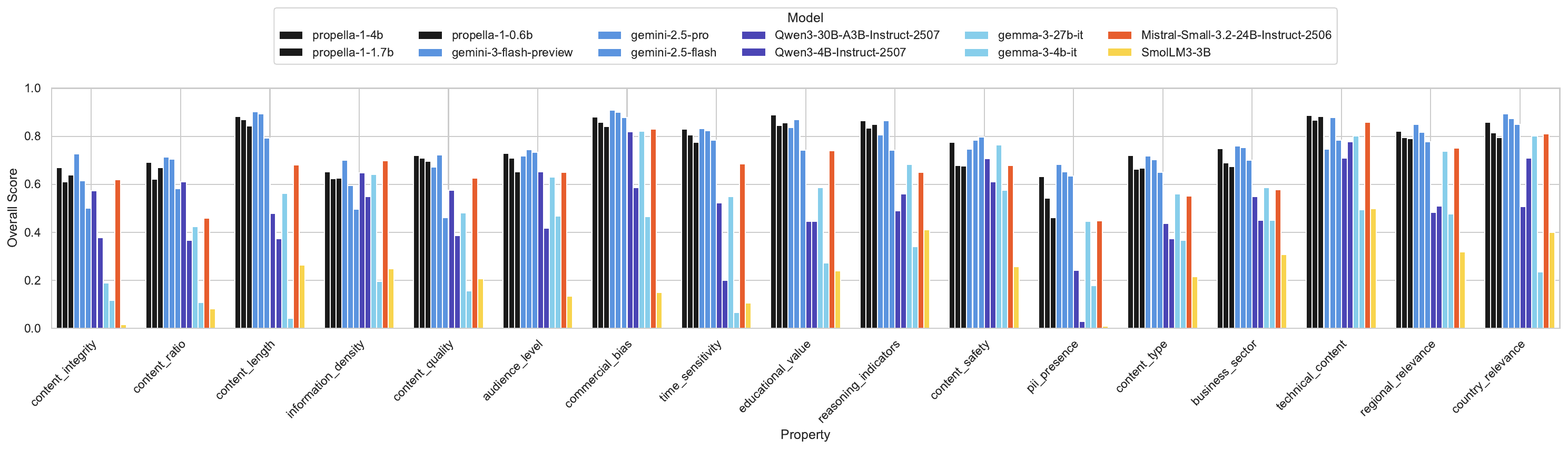}
    \caption{Per-property annotation agreement scores across all evaluated models and all 17 evaluated properties.}
    \label{fig:perproperty_full}
\end{figure}
\end{landscape}

\section{Inference Precision: fp8 vs. bf16}
\label{app:fp8}

\propella{} models are trained with fp8 precision and work well in both bf16 and fp8 inference modes.
As Table~\ref{tab:fp8} shows, fp8 inference preserves annotation quality, with negligible differences compared to bf16.

\begin{table}[H]
\centering
\caption{Effect of inference precision on annotation quality (overall score).}
\label{tab:fp8}
\begin{tabular}{@{}lccc@{}}
\toprule
\textbf{Model} & \textbf{bf16} & \textbf{fp8} & \textbf{Diff} \\
\midrule
propella-1-4b & 0.779 & 0.783 & +0.51\% \\
propella-1-1.7b & 0.737 & 0.731 & $-$0.81\% \\
\bottomrule
\end{tabular}
\end{table}

\pagebreak
\section{Complete Property Reference}
\label{app:properties}

Below is the complete annotation rubric for all 18 \propella{} properties.
For each property, we provide the full operational definition including enumerated values, decision criteria, borderline examples, decision trees, and language-specific guidelines.
The rubric is also released alongside the model weights.

\subsection{Core Content Properties}

\subsubsection{Content Integrity}

\textit{What we measure:} Completeness and technical quality of the content itself, regardless of navigation ratio.

\begin{description}
\item[\texttt{complete}] Full, intact content as intended.
\begin{itemize}
\item Content appears complete with proper beginning, middle, and end.
\item All essential elements present (introduction, body, conclusion where appropriate).
\item No obvious truncation or missing sections.
\item Example: Complete articles, full tutorials, intact documents.
\end{itemize}

\item[\texttt{mostly\_complete}] Minor elements missing but core content intact.
\begin{itemize}
\item Core content is complete but some secondary elements may be missing.
\item Minor truncation that does not affect main message.
\item Example: Article with truncated comments, missing sidebar content, partial author bio.
\end{itemize}

\item[\texttt{fragment}] Incomplete content, missing significant portions.
\begin{itemize}
\item Missing introduction, conclusion, or substantial middle sections.
\item Truncated mid-sentence or mid-paragraph.
\item Content feels incomplete or cut off.
\item Example: Search result snippets, article excerpts, broken crawls, partial downloads.
\end{itemize}

\item[\texttt{severely\_degraded}] Broken, unreadable, or corrupted content.
\begin{itemize}
\item Encoding errors, scrambled text, missing characters.
\item Severely malformed HTML rendering as gibberish.
\item Technical corruption making content unreadable.
\item Example: Garbled characters, completely broken formatting, corrupted files.
\end{itemize}
\end{description}

\paragraph{Key Decision Points.}
\begin{itemize}
\item \textbf{Content completeness}: Does the content feel like a complete unit of information?
\item \textbf{Technical integrity}: Is the content technically readable and properly formatted?
\item \textbf{Fragment vs.\ complete}: Independent of navigation; is the actual content complete?
\item \textbf{Degraded vs.\ fragment}: Degraded has technical issues; fragment is just incomplete.
\end{itemize}

\paragraph{Note.} Documents may end with a special \texttt{<content\_truncated>} tag, indicating upstream length-based truncation due to processing constraints. Content Integrity should not be penalized for this truncation signal; assess integrity based on the visible content's coherence and technical readability, ignoring the artificial cutoff.

\subsubsection{Content Ratio}

\textit{What we measure:} How much of the document is actual content vs.\ navigation, UI elements, and structural markup.

\begin{description}
\item[\texttt{complete\_content}] 90--100\% meaningful content.
\begin{itemize}
\item Full articles, papers, tutorials with minimal navigation.
\item Clean text with proper paragraphs and structure.
\item Example: A Wikipedia article, academic paper, complete blog post.
\end{itemize}

\item[\texttt{mostly\_content}] 70--89\% meaningful content.
\begin{itemize}
\item Complete documents with some navigation elements (header, footer, sidebar).
\item Minor UI elements that do not disrupt reading.
\item Example: News articles with standard website navigation.
\end{itemize}

\item[\texttt{mixed\_content}] 40--69\% meaningful content.
\begin{itemize}
\item Significant navigation mixed throughout content.
\item Multiple sidebars, ads, or UI elements interrupting text.
\item Example: E-commerce product pages with reviews mixed with purchase options.
\end{itemize}

\item[\texttt{mostly\_navigation}] 10--39\% meaningful content.
\begin{itemize}
\item Predominantly menus, links, headers, footers.
\item Content overwhelmed by structural elements.
\item Example: Site maps, navigation pages, heavily UI-focused pages.
\end{itemize}

\item[\texttt{minimal\_content}] 0--9\% meaningful content.
\begin{itemize}
\item Almost entirely navigation, UI elements, or structural markup.
\item Very little readable content present.
\item Example: Empty pages, pure navigation menus, error pages with minimal text.
\end{itemize}
\end{description}

\paragraph{Key Decision Points.}
\begin{itemize}
\item Focus on the ratio of readable text to navigation/UI elements.
\item Count only substantive content; ignore boilerplate and structural elements.
\item \textbf{Mixed vs.\ mostly\_navigation}: Can you read it as coherent content despite distractions?
\end{itemize}

\subsubsection{Content Length}

\textit{What we measure:} Amount of substantive content, ignoring navigation and boilerplate.

\begin{description}
\item[\texttt{substantial}] 2,000+ words of meaningful content.
\begin{itemize}
\item Long-form, comprehensive content that provides in-depth coverage of a topic.
\item Typically includes detailed analysis, multiple sections or chapters, extensive research, or thorough exploration of complex subjects.
\item Examples: White papers, research reports, e-books, long-form journalism.
\end{itemize}

\item[\texttt{moderate}] 500--2,000 words of meaningful content.
\begin{itemize}
\item Standard-length content that offers meaningful coverage while remaining focused and digestible.
\item Balances depth with accessibility; provides enough detail to be informative without overwhelming readers.
\item Examples: Typical blog posts, news articles, product reviews, how-to guides.
\end{itemize}

\item[\texttt{brief}] 100--500 words of meaningful content.
\begin{itemize}
\item Short, focused content that delivers key information quickly and efficiently.
\item Gets straight to the point while still providing value and context.
\item Examples: News briefs, product descriptions, FAQs, short blog posts.
\end{itemize}

\item[\texttt{minimal}] Under 100 words of meaningful content.
\begin{itemize}
\item Very short content that provides only essential information or serves as a quick reference.
\item Designed for rapid consumption or specific micro-purposes.
\item Examples: Social media posts, announcements, abstracts, snippets, navigation pages.
\end{itemize}
\end{description}

\paragraph{Measurement Tips.}
\begin{itemize}
\item \textbf{Count only readable content of value}: include article body and substantive headings/captions; exclude headers/footers, menus/sidebars, related links, share/consent UI, pagination, ads, and boilerplate.
\item Focus on substantive information, not filler words.
\item Complete thoughts matter more than exact word counts.
\item \textbf{Contextual adjustment}: Thresholds are guidelines and can be adjusted based on specific use cases. Academic contexts may shift ranges upward, while social media contexts may shift them downward.
\end{itemize}

\subsection{Content Classification}

\subsubsection{One-Sentence Description}

\textit{What we produce:} A very short, neutral description of what the document contains.

\begin{description}
\item[\texttt{one\_sentence\_description}] Ultra-short neutral description of the document.
\begin{itemize}
\item Exactly one sentence; target length $<$100 characters.
\item Focus on the main topic and, if useful, the document's function (e.g., tutorial, policy, news report, product page, navigation page).
\item Neutral, descriptive tone (no hype or marketing language).
\end{itemize}
\end{description}

\paragraph{To Avoid.}
\begin{itemize}
\item Boilerplate intros: ``This document\ldots'', ``This article\ldots'', ``In this guide\ldots''
\item Calls to action: ``Learn how to\ldots'', ``Discover\ldots'', ``Find out\ldots''
\item User-facing phrasing: ``You will learn\ldots'', ``How do I\ldots''
\item Non-essential details (dates, numbers) unless central to the topic.
\end{itemize}

\paragraph{Examples.}
\begin{itemize}
\item ``Beginner tutorial on React hooks and basic state management.''
\item ``News report on European Central Bank interest rate decisions.''
\item ``Internal policy for customer data retention and deletion.''
\item ``API reference for payment processing endpoints and error codes.''
\item ``Research paper analyzing housing price trends in major US cities.''
\item ``FAQ answering common questions about employee parental leave.''
\item ``Opinion essay arguing for stricter international climate change legislation.''
\end{itemize}

\paragraph{Examples for low-quality or problematic documents (still annotate).}
\begin{itemize}
\item ``Fragment of article discussing proposed changes to European data privacy laws.''
\item ``Keyword-stuffed promotional page about cheap car insurance quotes.''
\item ``Website navigation page listing links to product categories and help pages.''
\item ``Error page explaining that the requested resource could not be found.''
\item ``Affiliate landing page promoting multiple online casino bonus offers.''
\item ``Corrupted text with no identifiable topic or meaningful content.''
\end{itemize}

\subsubsection{Content Type}

\textit{What we measure:} The functional structure and purpose of content.

Content can be assigned multiple type labels if it genuinely serves multiple purposes. Choose all applicable types rather than forcing a single primary choice. Always output an array for this property, even if only one type applies.

\begin{description}
\item[\texttt{analytical}] In-depth analysis, research, and critical examination.
Provides detailed analysis or research on a topic; develops arguments, evaluates evidence, or presents findings.
Example: research analysis, investigative reports, academic articles, expert commentary.

\item[\texttt{instructional}] Teaching and how-to content.
Explicitly teaches skills, concepts, or procedures; step-by-step guidance or educational explanations.
Example: tutorials, how-to guides, educational content, training materials.

\item[\texttt{reference}] Lookup materials, definitions, specifications.
Designed for looking up specific information rather than reading through; often organized alphabetically, categorically, or as lists.
Example: dictionaries, encyclopedias, API references, product catalogs.

\item[\texttt{procedural}] Step-by-step processes and procedures.
Sequential instructions or workflows; process documentation with clear steps.
Example: recipes, installation guides, standard operating procedures, workflows.

\item[\texttt{qa\_structured}] Structured question-answer content.
Formal Q\&A format with clear questions and answers; often expert responses to specific questions.
Example: Stack Overflow, FAQ sections, structured Q\&A sites.

\item[\texttt{conversational}] Multi-party or turn-based dialogues (humans, bots, or both).
Casual or structured conversations between two or more participants; may include human--AI chats, forum threads, or comment chains.
Example: Reddit threads, forum discussions, support chats, assistant chat logs.

\item[\texttt{creative}] Entertainment, artistic, fictional content.
Primary purpose is entertainment or artistic expression; not primarily informational or instructional.
Example: short stories, poems, movie reviews, game content, fiction.

\item[\texttt{transactional}] Commercial, shopping, service-oriented.
Primary purpose is to facilitate a transaction or service; focuses on products, services, or business processes.
Example: product listings, service descriptions, checkout pages.

\item[\texttt{boilerplate}] Legal, policy, standard template text.
Standard legal or policy language; often repeated across multiple sites with minimal variation.
Example: terms of service, privacy policies, disclaimers, cookie banners, standard notices.

\item[\texttt{news\_report}] Straight reporting of events with minimal analysis.
Describes events or facts in a neutral, descriptive tone; time-bound news, updates, or reports.
Example: wire-service news articles, breaking-news updates.

\item[\texttt{opinion\_editorial}] Persuasive/opinionated commentary or editorials.
Expresses a stance or argument; aims to persuade; may cite evidence but prioritizes viewpoint.
Example: op-eds, opinion columns, personal essays with clear stance.

\item[\texttt{review\_critique}] Evaluative reviews of products, media, or services.
Provides judgments, ratings, or critiques; may include pros/cons, scoring systems.
Example: product reviews, film/book critiques, app store reviews (long-form).

\item[\texttt{technical\_documentation}] Manuals, API docs, developer guides, READMEs.
Primary goal is to instruct usage of software/hardware/APIs; includes reference sections, examples, parameters, version notes.
Example: API reference, library README, user manual.

\item[\texttt{specification\_standard}] Normative standards and formal specifications.
Defines requirements, must/shall language, compliance criteria; maintained by standards bodies or authoritative groups.
Example: RFCs, ISO standards, formal protocol specs.

\item[\texttt{legal\_document}] Statutes, case law, contracts, regulatory texts.
Binding or authoritative legal content; formal legal language and structure.
Example: court opinions, legislation, contracts, regulatory rules.

\item[\texttt{press\_release}] Organization-issued announcements and PR materials.
Promotional announcements framed as information; quotes from executives, product/service announcements.
Example: company press releases, launch announcements.

\item[\texttt{structured\_data}] Tables, datasets, indices, catalogs with minimal prose.
Predominantly tabular/listed data meant for lookup; minimal narrative or explanatory text.
Example: product catalogs, schedules, statistical tables.

\item[\texttt{source\_code}] Code listings as primary content.
Dominant content is program source code or scripts; may include lightweight comments or snippets without narrative.
Example: code files, gist-like pages, competitive programming solutions.
\end{description}

\paragraph{Multi-Type Examples.}
\begin{itemize}
\item Tutorial that analyzes different approaches $\rightarrow$ \texttt{[instructional, analytical]}
\item Educational reference manual $\rightarrow$ \texttt{[instructional, reference]}
\item Research paper with step-by-step methodology $\rightarrow$ \texttt{[analytical, procedural]}
\item Q\&A site with analytical responses $\rightarrow$ \texttt{[qa\_structured, analytical]}
\item API guide with examples $\rightarrow$ \texttt{[technical\_documentation, reference, instructional]}
\item RFC with rationale $\rightarrow$ \texttt{[specification\_standard, analytical]}
\item Film review with interview snippets $\rightarrow$ \texttt{[review\_critique, conversational]}
\item Helpdesk chat with an AI $\rightarrow$ \texttt{[conversational, transactional]}
\end{itemize}

\subsubsection{Business Sector}

\textit{What we measure:} Business sector(s) or industry domain(s) for training sector-specific LLMs.

Content can be assigned multiple sector labels if it genuinely spans multiple industries. Choose all applicable sectors rather than forcing a single primary choice or using ``other''. Always output an array.

\begin{description}
\item[\texttt{academic\_research}] Scholarly and research content (journal articles, conference papers, dissertations).
\item[\texttt{education\_sector}] Educational institutions and pedagogy (curricula, edtech, teaching resources).
\item[\texttt{technology\_software}] Software and information technology (development, platforms, IT services).
\item[\texttt{hardware\_electronics}] Hardware devices and electronics (semiconductors, embedded systems, datasheets).
\item[\texttt{healthcare\_medical}] Healthcare and medical sector (clinical practice, medical devices, public health).
\item[\texttt{pharmaceutical\_biotech}] Pharmaceutical and biotechnology (drug development, clinical trials, life sciences).
\item[\texttt{financial\_services}] Banking and financial services (investment, fintech, asset management).
\item[\texttt{legal\_services}] Legal sector and jurisprudence (law firms, compliance, litigation).
\item[\texttt{government\_public}] Government and public administration (agencies, policy, regulatory bodies).
\item[\texttt{manufacturing\_industrial}] Manufacturing and heavy industry (production, supply chain, industrial engineering).
\item[\texttt{mining\_resources}] Mining and natural resources (exploration, extraction, commodity operations).
\item[\texttt{chemicals\_materials}] Chemicals and advanced materials (petrochemicals, polymers, materials science).
\item[\texttt{energy\_utilities}] Energy and utilities (power generation, renewables, grid management).
\item[\texttt{retail\_commerce}] Retail and e-commerce (retail operations, merchandising, consumer experience).
\item[\texttt{wholesale\_distribution}] Wholesale trade and distribution (B2B, procurement, fulfillment).
\item[\texttt{real\_estate\_construction}] Real estate and construction (property development, architecture).
\item[\texttt{transportation\_logistics}] Transportation and logistics (shipping, freight, fleet management).
\item[\texttt{travel\_aviation}] Travel industry and commercial aviation (airlines, OTA platforms, IATA regulations).
\item[\texttt{automotive\_industry}] Automotive manufacturing and services (vehicle technology, electric vehicles).
\item[\texttt{telecommunications}] Telecommunications (network infrastructure, mobile services, 5G).
\item[\texttt{media\_entertainment}] Media and entertainment (film, television, music, streaming, publishing).
\item[\texttt{gaming\_industry}] Video games and interactive entertainment (development, esports, live ops).
\item[\texttt{gambling\_betting}] Gambling, betting, and online casinos (sportsbooks, affiliate content).
\item[\texttt{advertising\_marketing}] Advertising, marketing, and PR (brand strategy, campaign planning, martech).
\item[\texttt{hospitality\_tourism}] Hospitality and tourism (hotels, destination management, event planning).
\item[\texttt{food\_beverage\_hospitality}] Food \& beverage and restaurant operations (menu engineering, food safety).
\item[\texttt{agriculture\_food}] Agriculture and food production (farming, agtech, food processing).
\item[\texttt{environmental\_services}] Environmental and sustainability services (ESG, remediation, impact assessments).
\item[\texttt{aerospace\_defense}] Aerospace and defense (aircraft manufacturing, space technology, military systems).
\item[\texttt{insurance\_industry}] Insurance sector (actuarial science, underwriting, risk assessment).
\item[\texttt{nonprofit\_ngo}] Nonprofit and NGO sector (charitable organizations, international development).
\item[\texttt{consulting\_professional}] Professional services and consulting (management consulting, business advisory).
\item[\texttt{human\_resources}] HR and people operations (talent acquisition, workforce planning, L\&D).
\item[\texttt{security\_cyber}] Security and cybersecurity (threat intelligence, incident response, compliance).
\item[\texttt{consumer\_goods}] Consumer products and CPG (household products, personal care, FMCG).
\item[\texttt{general\_interest}] General audience content (broad topics, cross-sector, lifestyle).
\item[\texttt{other}] Highly specialized or unclassifiable niches.
\end{description}

\paragraph{Multi-Sector Examples.}
\begin{itemize}
\item Medical device regulations $\rightarrow$ \texttt{healthcare\_medical} + \texttt{pharmaceutical\_biotech} + \texttt{government\_public}
\item Fintech software documentation $\rightarrow$ \texttt{financial\_services} + \texttt{technology\_software}
\item Agricultural biotechnology research $\rightarrow$ \texttt{agriculture\_food} + \texttt{pharmaceutical\_biotech}
\end{itemize}

\subsubsection{Technical Content}

\textit{What we measure:} Type and intensity of specialized technical knowledge.

Content can be assigned multiple technical content labels if it genuinely combines multiple technical domains. Always output an array.

\begin{description}
\item[\texttt{code\_heavy}] Significant programming content.
Multiple code examples, algorithms, or implementations; software development focus.
Example: programming tutorials, API documentation, software guides.

\item[\texttt{math\_heavy}] Substantial mathematical content.
Mathematical equations, proofs, or statistical analysis; quantitative analysis and mathematical reasoning.
Example: mathematical papers, statistical analysis, quantitative research.

\item[\texttt{scientific}] Research and scientific methodology content.
Scientific research findings, experimental data; peer-reviewed research content.
Example: research papers, scientific studies, experimental reports.

\item[\texttt{data\_heavy}] Substantial datasets, tables, and data analysis.
Contains significant data tables, charts, or datasets; focus on data interpretation.
Example: research data, statistical reports, survey results.

\item[\texttt{engineering}] Engineering and applied technical content.
Engineering design, systems, and applied technical solutions; non-software engineering disciplines.
Example: mechanical engineering, civil engineering, technical specifications, design documents.

\item[\texttt{basic\_technical}] Some technical elements but not dominant.
Light technical content mixed with general explanations; technical concepts explained for general audience.
Example: technology articles for general audience, basic technical explanations.

\item[\texttt{non\_technical}] No significant technical content.
General audience content without specialized technical knowledge; no programming, mathematical, engineering, or scientific focus.
Example: general articles, humanities content, basic informational content.
\end{description}

\paragraph{Multi-Technical Examples.}
\begin{itemize}
\item Data science tutorial with code examples $\rightarrow$ \texttt{[code\_heavy, math\_heavy, data\_heavy]}
\item Engineering research with statistical analysis $\rightarrow$ \texttt{[engineering, scientific, data\_heavy]}
\item Computational biology paper $\rightarrow$ \texttt{[code\_heavy, scientific]}
\end{itemize}

\subsection{Quality and Value Assessment}

\subsubsection{Content Quality}

\textit{What we measure:} Overall quality of content considering writing excellence, substantive value, and presentation quality regardless of authorship origin.

\begin{description}
\item[\texttt{excellent}] Outstanding quality across all dimensions.
\begin{itemize}
\item Sophisticated writing with varied sentence structures and engaging style.
\item Rich, appropriate vocabulary with error-free grammar and punctuation.
\item High substantive value with clear insights or information.
\item Professional presentation and formatting; natural flow and logical organization.
\item Example: high-quality publications, expert analyses, polished educational content.
\end{itemize}

\item[\texttt{good}] High quality with minor imperfections.
\begin{itemize}
\item Grammatically correct with proper sentence structure; appropriate vocabulary and tone.
\item Solid substantive value and clear information; good organization and readable flow.
\item Only occasional minor issues (1--2 typos per section).
\item Example: quality journalism, professional websites, well-written blog posts.
\end{itemize}

\item[\texttt{adequate}] Acceptable quality for most purposes.
\begin{itemize}
\item Generally clear and understandable writing; some grammatical errors but meaning remains clear.
\item Reasonable substantive value though may lack depth; basic organization present.
\item Example: casual blogs, user reviews, basic informational content, simple guides.
\end{itemize}

\item[\texttt{poor}] Significant quality issues impacting utility.
\begin{itemize}
\item Multiple errors affecting comprehension or credibility.
\item Unclear expression, confusing organization, or awkward phrasing.
\item Limited substantive value or questionable information.
\item Example: low-quality web content, poorly edited materials, confusing instructions.
\end{itemize}

\item[\texttt{unacceptable}] Quality too low for productive use.
\begin{itemize}
\item Severely impaired communication with major errors; incoherent, nonsensical, or corrupted content.
\item No reliable substantive value; broken formatting or technical corruption.
\item Example: corrupted text, severe translation errors, spam content, SEO content.
\end{itemize}
\end{description}

\paragraph{Quality Assessment Guidelines.}
\begin{itemize}
\item \textbf{Comprehension}: Can the intended message be clearly understood?
\item \textbf{Substantive value}: Does the content provide useful information or insights?
\item \textbf{Technical presentation}: Is the content properly formatted and readable?
\item \textbf{Error impact}: Do errors significantly impede understanding or credibility?
\item \textbf{Professional standards}: Does the content meet basic standards for its intended purpose?
\end{itemize}

\paragraph{Language-Specific Quality Indicators.}
\begin{itemize}
\item For non-Latin scripts (Arabic, Chinese, Japanese): check for proper character encoding.
\item For agglutinative languages (Turkish, Finnish): adjust expectations for word count/density.
\item For languages with different formality levels (Japanese, Korean): assess appropriate register.
\item Mixed-language documents: evaluate code-switching quality and appropriateness.
\end{itemize}

\subsubsection{Information Density}

\textit{What we measure:} Ratio of valuable information to redundancy, padding, and repetition.

\begin{description}
\item[\texttt{dense}] Efficient, information-packed content.
Every sentence adds new information or insight; minimal redundancy or unnecessary elaboration; little to no repetition.
Example: technical specifications, concise academic writing, quality reference material.

\item[\texttt{adequate}] Good information content with reasonable elaboration.
Most content adds value with some acceptable elaboration; minimal repetition; good balance of information and explanation.
Example: well-written articles, good tutorials with examples.

\item[\texttt{moderate}] Mixed substantive content with noticeable padding.
Some valuable information mixed with unnecessary elaboration; noticeable repetition of key points.
Example: blog posts with some fluff, articles with repetitive conclusions.

\item[\texttt{thin}] Low information content with significant problems.
Much content does not add new information; high internal repetition and excessive redundancy; significant padding.
Example: SEO-optimized content, poorly edited writing.

\item[\texttt{empty}] Dominated by repetition and meaningless content.
Minimal actual information value; dominated by repetition and copy-paste artifacts; same ideas repeated without development.
Example: spam content, template-filled pages, keyword-stuffed articles.
\end{description}

\paragraph{Common Repetition Patterns to Watch For.}
\begin{itemize}
\item Same phrases repeated throughout (especially in SEO content).
\item Identical paragraphs or sections (copy-paste errors).
\item Circular reasoning (saying the same thing in different ways).
\item Template artifacts (repeated boilerplate mixed with content).
\end{itemize}

\subsubsection{Educational Value}

\textit{What we measure:} Potential for teaching, learning, and knowledge transfer.

\begin{description}
\item[\texttt{high}] Clear instructional design and learning objectives.
\begin{itemize}
\item Explicitly teaches concepts or skills; progressive skill building from basic to advanced.
\item Clear learning objectives and outcomes; comprehensive explanations with examples.
\item Example: quality tutorials, textbooks, structured courses, educational guides.
\end{itemize}

\item[\texttt{moderate}] Good instructional value with some learning potential.
\begin{itemize}
\item Some instructional elements present; explanations help build understanding.
\item Transferable knowledge to other contexts; good examples or illustrations.
\item Example: how-to articles, explanatory content, informative guides.
\end{itemize}

\item[\texttt{basic}] Limited educational content.
\begin{itemize}
\item Some explanations but not systematically instructional; basic explanations of concepts.
\item Limited learning potential or skill building.
\item Example: basic explanations, simple informational content.
\end{itemize}

\item[\texttt{minimal}] Little educational value.
\begin{itemize}
\item Primarily informational rather than instructional; no clear learning objectives.
\item Entertainment or commercial focus.
\item Example: entertainment content, basic news, commercial content.
\end{itemize}

\item[\texttt{none}] No educational content.
\begin{itemize}
\item No instructional value or learning potential; purely transactional, entertainment, or administrative.
\item Example: pure entertainment, transactions, legal boilerplate.
\end{itemize}
\end{description}

\paragraph{Disambiguation Tips.}
\begin{itemize}
\item Explanatory vs.\ educational: explanations alone $\neq$ educational design; require intent to teach plus scaffolding for Basic+.
\item Reference docs: typically Minimal; promote to Basic/Moderate when guided ``how-to'' segments or curated examples exist.
\item Reviews/op-eds: None/Minimal unless they include actionable how-to guidance designed for learning.
\end{itemize}

\paragraph{Automation Heuristics.}
\begin{itemize}
\item Keywords: Objectives/Outcomes, Lesson, Exercise/Quiz, Homework, Assessment, Syllabus, Module, Unit, Learning Goals.
\item Structure: numbered steps + prerequisites/requirements $\rightarrow$ Basic; add practice tasks/solutions $\rightarrow$ Moderate; syllabus/modules/assessments $\rightarrow$ High.
\item Signals of non-educational mix: heavy CTAs/ads or product pitches $\rightarrow$ cap at Minimal unless clear instructional scaffolding.
\end{itemize}

\paragraph{Quick Decision Tree.}
\begin{itemize}
\item Are there explicit learning goals or a syllabus? $\rightarrow$ High.
\item Else, are there step-by-step instructions with examples/exercises? $\rightarrow$ Moderate.
\item Else, are there explanatory sections intended to teach basics? $\rightarrow$ Basic.
\item Else, is there any minor instructional element? $\rightarrow$ Minimal.
\item Otherwise $\rightarrow$ None.
\end{itemize}

\paragraph{Borderline Examples.}
\begin{itemize}
\item API reference with examples but no guidance $\rightarrow$ Minimal to Basic (depending on clarity/examples).
\item Blog post explaining concept with analogies and one example $\rightarrow$ Basic.
\item Tutorial with tasks, checkpoints, and solutions $\rightarrow$ High.
\item Product documentation with ``Getting Started'' and ``How-To'' flows $\rightarrow$ Moderate.
\end{itemize}

\paragraph{Educational Indicators.}
\begin{itemize}
\item \textbf{Learning objectives}: Clear goals for what reader should learn.
\item \textbf{Skill progression}: Builds from basic to advanced concepts.
\item \textbf{Examples and practice}: Provides concrete examples or exercises.
\item \textbf{Knowledge transfer}: Concepts applicable beyond immediate context.
\end{itemize}

\subsubsection{Reasoning Indicators}

\textit{What we measure:} Presence and quality of logical reasoning, analysis, and explanatory content.

\begin{description}
\item[\texttt{analytical}] Complex reasoning and systematic analysis.
\begin{itemize}
\item Multi-step arguments with logical progression; cause-effect analysis and systematic thinking.
\item Considers multiple perspectives or variables; draws conclusions from evidence and reasoning.
\item Example: research analysis, complex problem-solving, systematic evaluations.
\end{itemize}

\item[\texttt{explanatory}] Clear explanations with logical flow.
\begin{itemize}
\item Explains how or why things work; shows cause-effect relationships clearly.
\item Educational reasoning that builds understanding; logical connections between concepts.
\item Example: good tutorials, educational content, how-to explanations.
\end{itemize}

\item[\texttt{basic\_reasoning}] Simple logical connections.
\begin{itemize}
\item Some logical connections between ideas; basic explanations of concepts or processes.
\item Elementary analytical thinking; simple cause-effect relationships.
\item Example: basic explanations, simple arguments, elementary analysis.
\end{itemize}

\item[\texttt{minimal}] Limited reasoning, mostly descriptive.
\begin{itemize}
\item Primarily describes what rather than why or how; few logical connections between ideas.
\item Mostly factual statements without analysis; little explanatory content.
\item Example: basic descriptions, simple factual content, minimal analysis.
\end{itemize}

\item[\texttt{none}] No clear reasoning present.
\begin{itemize}
\item Purely descriptive content; simple factual listing without connections.
\item Narrative content without analysis; no logical argumentation or explanation.
\item Example: simple lists, basic narratives, pure description.
\end{itemize}
\end{description}

\paragraph{Thinking-Trace Signals.}
\begin{itemize}
\item Stepwise structure: numbered steps in proofs/derivations/solutions; ``First\ldots{} therefore\ldots{} hence\ldots{} so\ldots''
\item Hypothesis and test: assumptions, intermediate results, counterexamples, sanity checks.
\item Tool- or method-calls: named algorithms, theorems, lemmas, or procedures invoked and justified.
\item Error analysis or reflection: ``we tried X, failed because Y, so we\ldots'', ``limitations,'' ``edge cases.''
\item Intermediate artifacts: scratch calculations, partial code reasoning, sub-problems and sub-claims.
\end{itemize}

\paragraph{Disambiguation Rules.}
\begin{itemize}
\item Explanatory vs.\ analytical: explanations tell how; analytical shows multi-step inference with evidence and intermediate claims.
\item Worked example vs.\ mere answer: worked examples expose steps and justification; mere answers without steps are not reasoning-rich.
\item Procedural vs.\ reasoning: procedural lists actions; reasoning links actions via logic, evidence, or constraints.
\end{itemize}

\paragraph{Automation Heuristics.}
\begin{itemize}
\item Lexical cues: because, therefore, thus, hence, suppose/assume, we conclude, by induction, lemma/theorem/proof, $O(n)$, hypothesis, counterexample.
\item Structure cues: presence of proof blocks, derivations (e.g., ``Proof.'', ``QED'', \TeX{} environments), multi-step numeric calculations.
\item Program reasoning: code comments like \texttt{// invariant}, \texttt{// complexity}, pre/post-conditions, test reasoning.
\item Thresholding: count reasoning cues per 1k tokens; with $\geq$2 structural cues or $\geq$5 lexical cues $\rightarrow$ at least explanatory; proofs/derivations $\rightarrow$ analytical.
\end{itemize}

\paragraph{Quick Decision Tree.}
\begin{itemize}
\item Is there a proof/derivation or multi-step argument with intermediate claims? $\rightarrow$ Analytical.
\item Else, does it explain why/how with cause-effect and logical links? $\rightarrow$ Explanatory.
\item Else, are there simple logical connections or one-step justifications? $\rightarrow$ Basic reasoning.
\item Else, does it mostly describe without connecting ideas? $\rightarrow$ Minimal/None.
\end{itemize}

\paragraph{Borderline Examples.}
\begin{itemize}
\item Answer-only solutions (final numeric result without steps) $\rightarrow$ Minimal.
\item Step-by-step math solution with intermediate equations $\rightarrow$ Analytical.
\item ``How it works'' article connecting 2--3 causal steps without data $\rightarrow$ Explanatory.
\item Troubleshooting log with attempts and justifications $\rightarrow$ Analytical if causal chain is explicit; otherwise Explanatory.
\end{itemize}

\paragraph{Key Reasoning Patterns to Identify.}
\begin{itemize}
\item \textbf{Cause-effect}: ``Because X, therefore Y.''
\item \textbf{Problem-solution}: Identifies problems and proposes solutions.
\item \textbf{Comparison}: Analyzes similarities and differences.
\item \textbf{Logical progression}: Ideas build on previous ideas.
\item \textbf{Evidence-based conclusions}: Draws conclusions from presented evidence.
\end{itemize}

\subsection{Audience and Purpose}

\subsubsection{Audience Level}

\textit{What we measure:} Intended sophistication level and background knowledge assumptions of the target audience.

\begin{description}
\item[\texttt{expert}] Highly specialized professional/academic content.
\begin{itemize}
\item Assumes deep domain expertise and advanced training; uses technical terminology without explanation.
\item Content for practitioners actively working in specialized fields.
\item Example: climate modeling methodology in Nature Climate Change, research papers, technical specifications.
\end{itemize}

\item[\texttt{advanced}] Educated adult audience with analytical skills.
\begin{itemize}
\item Assumes higher education and critical thinking ability; explains specialized concepts but uses sophisticated language.
\item Intellectually challenging but accessible to educated generalists.
\item Example: complex climate change analysis in The Atlantic, quality journalism, policy analysis.
\end{itemize}

\item[\texttt{general}] General adult audience.
\begin{itemize}
\item Accessible to most educated adults without specialized background; explains technical concepts when introduced.
\item Uses clear language while maintaining intellectual substance.
\item Example: quality journalism, general interest articles, accessible explanations of complex topics.
\end{itemize}

\item[\texttt{beginner}] Introductory level with minimal prerequisites.
\begin{itemize}
\item Explains basic concepts and terminology; builds up from fundamental principles.
\item Assumes minimal prior knowledge of the subject area.
\item Example: introductory tutorials, beginner guides, basic explanations, getting-started content.
\end{itemize}

\item[\texttt{youth}] Targeted at teenagers and young adults (ages 13--19).
\begin{itemize}
\item Age-appropriate complexity with contemporary cultural references.
\item Sophisticated enough for developing critical thinking but accessible.
\item Example: high school educational content, young adult literature, college prep materials.
\end{itemize}

\item[\texttt{children}] Designed specifically for children.
\begin{itemize}
\item Simple language and concepts appropriate for young readers.
\item Educational content designed for elementary/middle school levels.
\item Example: children's educational content, elementary school materials, simple explanations for young learners.
\end{itemize}
\end{description}

\paragraph{Assessment Guidelines.}
\begin{itemize}
\item \textbf{Professional context}: Is this content designed for workplace use vs.\ general learning?
\item \textbf{Terminology density}: How much specialized vocabulary is used without explanation?
\item \textbf{Concept complexity}: How sophisticated are the ideas and their development?
\item \textbf{Background assumptions}: What education level and domain knowledge does the author assume?
\end{itemize}

\paragraph{Cross-Linguistic Considerations.}
\begin{itemize}
\item Expert terminology density varies by language (German allows more compound terms).
\item Formality markers differ across cultures.
\item Educational level assumptions vary by country's education system.
\item Age-appropriate content differs across cultures.
\end{itemize}

\subsubsection{Commercial Bias}

\textit{What we measure:} How much commercial interests influence the objectivity and informational value of content.

\begin{description}
\item[\texttt{none}] No commercial influence detected.
Objective, informational presentation; no promotional language or commercial agenda; focus purely on informing or educating.
Example: academic papers, objective journalism, educational content.

\item[\texttt{minimal}] Slight commercial context but maintains objectivity.
May mention products/services but in informational context; maintains balanced, objective tone.
Example: product reviews with balanced analysis, informational articles mentioning relevant products.

\item[\texttt{moderate}] Some commercial influence on content.
Mix of informational and promotional content; some promotional language but still provides useful information.
Example: company blogs with useful information, sponsored content with actual value.

\item[\texttt{heavy}] Strong commercial bias throughout.
Primarily promotional with some informational elements; heavy use of marketing language and persuasive techniques.
Example: marketing articles disguised as information, heavily biased product comparisons.

\item[\texttt{pure\_marketing}] Entirely commercial/promotional content.
No genuine informational value beyond promotion; pure marketing copy or advertising material; designed solely to drive sales or conversions.
Example: sales pages, pure advertising copy, promotional brochures.
\end{description}

\paragraph{Key Indicators.}
\begin{itemize}
\item \textbf{Language tone}: Objective vs.\ promotional language.
\item \textbf{Primary purpose}: Inform vs.\ persuade/sell.
\item \textbf{Balance}: Are alternatives/drawbacks mentioned?
\item \textbf{Call-to-action}: Subtle information vs.\ obvious sales pitch.
\end{itemize}

\subsubsection{Time-Sensitivity}

\textit{What we measure:} How time-sensitive the content is, i.e., whether its value degrades over time or remains stable.

\begin{description}
\item[\texttt{evergreen}] Content remains valuable indefinitely.
Fundamental concepts, principles, theories; historical information and established facts; skills and techniques that do not change.
Example: mathematical proofs, language grammar guides, classical literature analysis, basic cooking techniques.

\item[\texttt{slowly\_changing}] Content remains valuable for years.
Best practices that evolve slowly; technical content that updates every few years; cultural and social topics with gradual change.
Example: programming language tutorials, academic textbooks, industry standards, educational curricula.

\item[\texttt{regularly\_updating}] Content valuable for months to a year.
Industry trends and market analysis; technology reviews and comparisons; policy discussions and current research.
Example: software framework guides, business strategies, product reviews, research summaries.

\item[\texttt{time\_sensitive}] Content value degrades quickly.
News and current events; time-bound information (prices, schedules, availability); temporary situations or short-term trends.
Example: stock prices, weather reports, breaking news, event announcements, sales/promotions.
\end{description}

\paragraph{Key Decision Points.}
\begin{itemize}
\item \textbf{Core question}: If someone reads this in 2 years, will it still be valuable?
\item \textbf{Update frequency}: How often does this type of information typically change?
\item \textbf{Temporal references}: Does the content heavily reference ``now,'' ``recently,'' ``currently''?
\item \textbf{Subject matter stability}: Is this about unchanging principles or evolving situations?
\end{itemize}

\subsection{Safety and Compliance}

\subsubsection{Content Safety}

\textit{What we measure:} Presence of inappropriate, harmful, or legally problematic content.

\begin{description}
\item[\texttt{safe}] Appropriate for all contexts.
No concerning content of any type; professional, appropriate language throughout; suitable for general audiences including workplace settings.

\item[\texttt{mild\_concerns}] Minor issues that do not constitute major problems.
Occasional mild profanity in context; brief mentions of sensitive topics handled appropriately.
Example: historical discussions of sensitive topics, professional content with mild language.

\item[\texttt{nsfw}] Not safe for work or general audiences.
Explicit sexual content or graphic descriptions; adult themes requiring content warnings; graphic violence or disturbing imagery descriptions.
Example: adult content, graphic medical descriptions, explicit violence.

\item[\texttt{harmful}] Potentially harmful content requiring careful handling.
Content promoting dangerous activities or self-harm; hate speech targeting individuals or groups; violent content glorifying harm to others.
Example: self-harm content, hate speech, dangerous ``how-to'' guides.

\item[\texttt{illegal}] Illegal content requiring immediate rejection.
Content promoting clearly illegal activities; material that violates laws in major jurisdictions.
Example: terrorist content, child exploitation.
\end{description}

\paragraph{Safety Assessment Guidelines.}
\begin{itemize}
\item \textbf{Context matters}: Medical/educational discussions of sensitive topics may be appropriate.
\item \textbf{Intent matters}: Discussing harmful topics for educational purposes vs.\ promoting them.
\item \textbf{Audience consideration}: Content appropriate for experts may not be safe for general audiences.
\end{itemize}

\subsubsection{PII Presence}

\textit{What we measure:} Whether the content contains personally identifiable information that could identify private individuals.

\begin{description}
\item[\texttt{no\_pii}] No personal information detected.
\begin{itemize}
\item No names of private individuals; no contact information (emails, phones, addresses); no identification numbers.
\item Public figures and officials mentioned by name are acceptable.
\item Example: news articles about politicians, technical documentation, general information.
\end{itemize}

\item[\texttt{contains\_pii}] Contains potentially identifiable information.
\begin{itemize}
\item Names of private individuals (non-public figures).
\item Email addresses, phone numbers, physical addresses.
\item ID numbers (SSN, passport, driver's license, employee IDs).
\item Medical information about identifiable individuals; financial account information.
\item Example: personal blogs with full names, leaked databases, medical case studies with identifying info.
\end{itemize}
\end{description}

\paragraph{Key Decision Points.}
\begin{itemize}
\item \textbf{Public vs.\ Private figures}: Politicians, celebrities, CEOs = public (no PII flag); private citizens = PII.
\item \textbf{Context matters}: Academic paper authors and their institutional emails = typically no PII; personal emails in forums = PII.
\item \textbf{Aggregated vs.\ Individual}: Statistical data = no PII; individual records = PII.
\end{itemize}

\subsection{Geographic Relevance}

\subsubsection{Regional Relevance}

\textit{What we measure:} Primary regional, cultural, or geopolitical sphere(s) that the content relates to, regardless of language used.

Content can be assigned multiple regional labels if it genuinely spans multiple regions. Choose all applicable regions rather than forcing a single primary choice. Always output an array.

\begin{description}
\item[\texttt{european}] European context (EU and broader Europe).
Content about European countries, EU policies, or pan-European topics; European cultural perspectives, social systems, or business practices. Includes EU member states, UK, Switzerland, Norway, Balkans, etc.
Example: GDPR compliance, European Parliament elections, Schengen area travel, European football leagues.

\item[\texttt{north\_american}] North American context.
Content about US, Canada, or Mexico; North American cultural perspectives, USMCA/NAFTA region topics.
Example: FDA regulations, Silicon Valley tech, NHL, US constitutional law, Canadian healthcare.

\item[\texttt{east\_asian}] East Asian context.
Content about China, Japan, Korea (North/South), Taiwan, Mongolia; East Asian cultural perspectives, Confucian-influenced societies.
Example: Gaokao exams, K-pop, Shenzhen tech hub, Japanese work culture, Taiwan semiconductor industry.

\item[\texttt{south\_asian}] South Asian context.
Content about India, Pakistan, Bangladesh, Sri Lanka, Nepal, Bhutan, Afghanistan, Maldives; South Asian cultural perspectives.
Example: IIT entrance exams, Bollywood, cricket leagues, monsoon impacts.

\item[\texttt{southeast\_asian}] Southeast Asian context.
Content about ASEAN countries (Indonesia, Thailand, Vietnam, Philippines, Malaysia, Singapore, etc.); Southeast Asian regional perspectives.
Example: ASEAN economic community, Indonesian elections, Singapore financial sector, Thai tourism.

\item[\texttt{middle\_eastern}] Middle Eastern and North African context.
Content about Arab states, Iran, Turkey, Israel, North Africa (MENA region); Islamic finance, regional conflicts.
Example: Gulf Cooperation Council, OPEC decisions, Middle East peace process, Islamic banking.

\item[\texttt{sub\_saharan\_african}] Sub-Saharan African context.
Content about African countries south of the Sahara; African Union topics, sub-Saharan development issues.
Example: M-Pesa mobile banking, African Union policies, safari tourism, ubuntu philosophy.

\item[\texttt{latin\_american}] Latin American context.
Content about Central and South America, Caribbean; regional integration (Mercosur, etc.).
Example: Mercosur trade, telenovelas, Amazon rainforest, Latin American revolutions.

\item[\texttt{oceanian}] Oceanian context.
Content about Australia, New Zealand, Pacific Island nations; Oceanian perspectives, Pacific regional issues.
Example: ANZAC relations, Pacific Island climate change, Australian mining, M\={a}ori culture.

\item[\texttt{central\_asian}] Central Asian context.
Content about Kazakhstan, Uzbekistan, Turkmenistan, Tajikistan, Kyrgyzstan; post-Soviet regional dynamics, Silk Road region.
Example: Silk Road initiatives, Caspian Sea resources, post-Soviet transitions.

\item[\texttt{russian\_sphere}] Russian/Post-Soviet context.
Content about Russia, Belarus, and strong Russian influence areas; CIS (Commonwealth of Independent States) topics.
Example: Russian federal politics, CIS integration, post-Soviet economic transitions.

\item[\texttt{global}] Genuinely international or universal.
Content with truly global scope or application; international organizations, worldwide phenomena, global comparisons.
Example: UN reports, climate change (global perspective), international standards, pandemic response.

\item[\texttt{culturally\_neutral}] No clear regional focus.
Abstract, theoretical, or technical content without regional markers; universal scientific, mathematical, or philosophical content.
Example: mathematical proofs, chemical formulas, abstract philosophy, programming concepts.

\item[\texttt{indeterminate}] Cannot determine regional relevance.
Insufficient content to identify regional focus; mixed or contradictory regional signals; fragment or corrupted content.
Example: technical specifications without context, isolated data tables.
\end{description}

\paragraph{Multi-Regional Examples.}
\begin{itemize}
\item EU-China trade relations $\rightarrow$ \texttt{[european, east\_asian]}
\item NAFTA/USMCA impact on Mexican agriculture $\rightarrow$ \texttt{[north\_american, latin\_american]}
\item Indian diaspora in the Gulf states $\rightarrow$ \texttt{[south\_asian, middle\_eastern]}
\item Comparative study of healthcare systems globally $\rightarrow$ \texttt{[global]}
\end{itemize}

\paragraph{Regional Identification Guidelines.}

\textbf{Primary indicators:}
\begin{itemize}
\item \textbf{Geographic references}: Countries, cities, regions, landmarks mentioned.
\item \textbf{Institutional references}: Governments, companies, universities, organizations specific to region.
\item \textbf{Cultural markers}: Holidays, customs, cultural phenomena, sports, entertainment.
\item \textbf{Political/economic systems}: References to regional political structures, economic blocs.
\item \textbf{Legal/regulatory frameworks}: Region-specific laws, regulations, standards.
\item \textbf{Language context}: While not determinative, language can provide regional hints.
\end{itemize}

\textbf{Important distinctions:}
\begin{itemize}
\item \textbf{Language $\neq$ Region}: Spanish content about Asian markets = \texttt{[east\_asian]}, not \texttt{[latin\_american]}.
\item \textbf{Company origin vs.\ topic}: Apple (US company) operating in India = consider actual content focus.
\item \textbf{Historical vs.\ current}: Historical content about ancient Rome = \texttt{[european]} if discussing modern implications.
\item \textbf{Diaspora content}: Content about diaspora communities should include both origin and current regions.
\end{itemize}

\textbf{Quality checks:}
\begin{itemize}
\item If content is in a non-English language but discusses global topics $\rightarrow$ still mark as \texttt{[global]}.
\item If content compares multiple regions $\rightarrow$ mark all regions discussed substantially.
\item If content is about a specific place but has universal applications $\rightarrow$ consider both regional and global tags.
\end{itemize}

\subsubsection{Country Relevance}

\textit{What we measure:} Which specific country or countries (if any) the content is relevant to, globally.

Always output an array of country names for this property (even when only a single country applies). Use standard country names from any region worldwide (e.g., ``germany'', ``france'', ``united\_states'', ``united\_kingdom'', ``china'', ``japan'', ``brazil'', ``india'', ``south\_africa'', ``australia'', ``canada'', etc.). The array may also contain the special values \texttt{supranational} or \texttt{none}.

\begin{description}
\item[\texttt{\{COUNTRY\_NAME\}}] Content specifically relevant to a single country.
\begin{itemize}
\item Content explicitly about that country's politics, culture, institutions, or regulations.
\item Content written from that country's cultural perspective.
\item Content about that country's cities, companies, institutions, or country-specific topics.
\item Only use country names listed in ISO-3166. Use ``united\_kingdom'' instead of ``england'', ``wales'', etc.
\item Example: For ``germany'' $\rightarrow$ German election coverage, Bundesliga content, German legal analysis.
\item Example: For ``japan'' $\rightarrow$ Japanese politics, J-League content, Japanese cultural analysis.
\end{itemize}

\item[\texttt{supranational}] Content focused on supranational entities or regions.
\begin{itemize}
\item International organizations, regional blocs, global institutions.
\item Content about supranational policies, international organizations, global governance.
\item Example: UN resolutions, NATO discussions, EU policy analysis, ASEAN agreements, WTO trade rules.
\end{itemize}

\item[\texttt{none}] Content not specifically relevant to any country.
\begin{itemize}
\item Abstract, theoretical, or universal content without geographic specificity.
\item Technical/scientific content that applies globally without country focus.
\item Example: mathematical proofs, universal scientific principles, abstract philosophical discussions.
\end{itemize}
\end{description}

\paragraph{Country Identification Criteria.}
\begin{itemize}
\item \textbf{Political content}: Elections, government policies, political parties, political figures specific to the country.
\item \textbf{Cultural content}: National traditions, cultural phenomena, historical events specific to the country.
\item \textbf{Institutional references}: Government bodies, national companies, universities specific to the country.
\item \textbf{Geographic focus}: Cities, regions, landmarks within the country as primary subjects.
\item \textbf{Legal/regulatory}: Laws, regulations, legal frameworks specific to the country.
\item \textbf{Economic content}: National economic policies, country-specific market analysis.
\item \textbf{Sports/media}: National sports leagues, national teams, country-specific media outlets.
\item \textbf{Social issues}: Social policies, demographic topics, social movements specific to the country.
\end{itemize}

\end{document}

%% file: math_commands.tex
\usepackage{amsmath,amsfonts,bm}

\def\eqref#1{equation~\ref{#1}}

\def\1{\bm{1}}

\DeclareMathAlphabet{\mathsfit}{\encodingdefault}{\sfdefault}{m}{sl}
\SetMathAlphabet{\mathsfit}{bold}{\encodingdefault}{\sfdefault}{bx}{n}

